\documentclass[fleqn,10pt]{wlscirep}
\usepackage{lipsum} 
\usepackage{bm}
\usepackage{times}
\usepackage{helvet}
\usepackage{courier}
\usepackage{times}
\usepackage{latexsym}
\usepackage{amsmath}
\usepackage{algorithm}
\usepackage{algpseudocode}
\usepackage{url}
\usepackage{multirow}
\usepackage{enumitem}
\usepackage{array}
\usepackage[symbol]{footmisc}
\usepackage{threeparttable}
\usepackage{graphicx}
\usepackage{dblfloatfix} 
\usepackage{caption}
\usepackage[labelformat=simple]{subcaption}

\usepackage{adjustbox}
\usepackage{multirow}
\usepackage{hyperref}
\usepackage{xcolor}
\usepackage{float}
\usepackage{longtable}

\usepackage{amsmath}
\usepackage{amssymb}
\usepackage{subcaption}
\usepackage{tabularx}
\usepackage{makecell}
\usepackage{mathtools}

\newcolumntype{P}[1]{>{\centering\arraybackslash}p{#1}}

\title{SynthEHR-Eviction: Enhancing Eviction SDoH Detection with LLM-Augmented Synthetic EHR Data}

\author[1, 2+]{Zonghai Yao, MSc}
\author[2+]{Youxia Zhao, MSc}
\author[1, 2]{Avijit Mitra, MSc}
\author[3]{David A. Levy, MD}
\author[1]{Emily Druhl, MPH}
\author[4, 5, 6]{Jack Tsai, PhD, MSCP}
\author[1, 2, 3, 7*]{Hong Yu, PhD}

\affil[1]{Center for Healthcare Organization and Implementation Research, VA Bedford Health Care, MA, USA}
\affil[2]{Manning College of Information and Computer Sciences, UMass Amherst, MA, USA}
\affil[3]{Department of Medicine, University of Massachusetts Medical School, Worcester, MA, USA}
\affil[4]{National Center on Homelessness among Veterans, VA Homeless Programs Office, Washington, DC, USA}
\affil[5]{School of Public Health, University of Texas Health Science Center at Houston, Houston, TX, USA}
\affil[6]{Department of Psychiatry, Yale University School of Medicine, New Haven, CT, USA}
\affil[7]{Miner School of Computer and Information Sciences, UMass Lowell, MA, USA}

\affil[*]{Corresponding author: Hong Yu (Hong\_Yu@uml.edu)}

\affil[+]{these authors contributed equally to this work}

\begin{abstract}

Eviction is a significant yet understudied social determinants of health (SDoH), linked to housing instability, unemployment, and mental health. 
While eviction appears in unstructured electronic health records (EHRs), it is rarely coded in structured fields, limiting downstream applications.
We introduce \textit{SynthEHR-Eviction}, a scalable pipeline combining LLMs, human-in-the-loop annotation, and automated prompt optimization (APO) to extract eviction statuses from clinical notes.
Using this pipeline, we created the largest public eviction-related SDoH dataset to date, comprising 14 fine-grained categories.
Fine-tuned LLMs (e.g., Qwen2.5, LLaMA3) trained on SynthEHR-Eviction achieved Macro-F1 scores of 88.8\% (eviction) and 90.3\% (other SDoH) on human validated data, outperforming GPT-4o-APO (87.8\%, 87.3\%), GPT-4o-mini-APO (69.1\%, 78.1\%), and BioBERT (60.7\%, 68.3\%), while enabling cost-effective deployment across various model sizes.
The pipeline reduces annotation effort by over 80\%, accelerates dataset creation, enables scalable eviction detection, and generalizes to other information extraction tasks.

\end{abstract}
\begin{document}

\maketitle

\section{Introduction}
\label{Intro}

In modern healthcare systems, advancing individualized, whole-person care and promoting access to comprehensive healthcare and social services have become key goals of global healthcare movements~\cite{nccih_wholeperson,bokhour2020transforming}. 
Beyond clinical indicators, environmental and social actors can also greatly influence health outcomes. 
The World Health Organization (WHO) has recognized the wide ranging influence of social determinants of health (SDoH) as "the conditions in which people are born, grow, live, work, and age,"~\cite{who_sdh} and notes that they contribute to as much as 80\%–90\% of health outcomes~\cite{hood2016county}.
Accurately and systematically integrating SDoH into clinical decision-making is a critical step toward achieving personalized care, promoting health and well-being, and enhancing the effectiveness of public health interventions~\cite{craig2021leveraging,rangachari2025impact}.
For example, individuals who are unstably housed may live in precarious, unsafe conditions that can affect their physical and mental health~\cite{tsai2021longitudinal,chen2022association}.
Lack of transportation can affect access to healthcare and daily necessities, such as healthy foods~\cite{tsai2024transportation,losada2024understanding}.

Yet, in current electronic health records (EHRs), SDoH information is often embedded within unstructured free-text clinical notes, lacking systematic, structured documentation~\cite{guevara2024large}. 
This results in "information silos", where crucial social risk factors are underutilized in real-world healthcare delivery, limiting clinicians’ ability to assess patients' broader health risks and restricting healthcare institutions and policymakers from designing proactive, targeted intervention strategies~\cite{wikipedia_eviction}.

Among various categories of SDoH, eviction is a highly impactful but long-overlooked factor affecting patient health.
Eviction can trigger a chain reaction of adverse outcomes, including housing instability, unemployment, homelessness, mental health issues, and long-term poverty~\cite{Desmond2015,Desmond2016}. 
Its impact extends beyond individuals, affecting entire communities and exacerbating social inequality~\cite{Tsai2019}. 
Studies have shown that marginalized populations—such as racial minorities, women, and low-income renters—face a significantly higher risk of eviction~\cite{Hepburn2020}. 
Particularly during the Coronavirus Disease-2019 (COVID-19) pandemic, eviction was directly associated with over 430,000 additional cases and more than 10,000 deaths in the United States, prompting the CDC to enact a federal eviction moratorium between 2020 and 2021~\cite{tsai2024eviction}. 
Despite its profound public health implications, the current healthcare system’s ability to identify and address eviction-related risks remained severely limited.
Mainstream public EHR datasets, such as MIMIC-III~\cite{johnson2016mimic} and MIMIC-IV~\cite{johnson2020mimic}, contain extremely sparse documentation of eviction-related information, with relevant terms appearing in fewer than 1 in 1,000 records.
Furthermore, standardized coding systems, such as relying on the International Classification of Diseases, Version 10 (ICD-10), lack specific codes for eviction, contributing to fragmented and underutilized eviction data in clinical practice~\cite{cms_sdoz2023}.
Unlocking and systematically incorporating eviction-related SDoH data can transform healthcare delivery by enabling comprehensive health risk assessments, integrating social work, housing support, and psychological interventions into patient care workflows, and providing real-time, high-quality data to support targeted public health interventions, particularly benefiting marginalized populations and advancing accessible healthcare and social service systems.

In response to these challenges and transformative demands, we develop a novel, scalable information extraction (IE) pipeline tailored for eviction-related SDoH annotation.
IE has long been a central goal in clinical NLP~\cite{sivarajkumar2024clinical}, with traditional approaches relying on rule-based extraction~\cite{savova2010mayo} or small model fine-tuning methods~\cite{hahn2020medical} that are time-consuming, difficult to scale, and often institution-specific. 
Recent advances in large language models (LLMs) have shown promise in scaling IE development through synthetic data generation and prompt-based learning~\cite{hu2024information}. 
These techniques can significantly reduce annotation costs and improve generalization, yet their use in complex, underrepresented SDoH domains remains limited~\cite{guevara2024large,keloth2025social,patra2021sdohextraction}.
Eviction, as a high-impact but rarely modeled SDoH~\cite{yao2023eviction}, exemplifies the need for clinically grounded, low-cost annotation solutions.

We introduce \textit{SynthEHR-Eviction}—a scalable and modular pipeline that integrates LLM-based data augmentation, automated prompt optimization (APO)~\cite{ramnath2025systematic}, and human-in-the-loop (HITL) validation to support large-scale, consistent annotation. During the augmentation phase, expert feedback is incorporated to refine prompts and balance label distributions.
For annotation, we adopt the DSPy framework~\cite{khattab2024dspy}, which enables chain-of-thought (CoT) annotation, automated reasoning trace generation, and fine-grained label refinement.
The pipeline involves two iterative stages: 
First, the augmenter generates new examples for underrepresented categories; 
Second, the annotator verifies and labels the generated notes while producing structured reasoning. 
Compared to full manual rewriting and annotation, this workflow reduces human labor by over 80\% while maintaining high data quality and interpretability. 
Moreover, its modular structure generalizes to other SDoH domains such as food insecurity, utility shutoff, or intimate partner violence~\cite{magister2022teaching}.

Using this pipeline, we constructed the SynthEHR-Eviction dataset, the largest publicly available dataset of SDoH related to evictions to date. 
It covered 14 carefully defined SDoH categories, including fine-grained eviction statuses such as "Eviction Absent," "Eviction Pending," and "Mutual Rescission History," (see Table~\ref{tab:definition_table_evi}) as well as related ICD-10 Z.59 categories like housing instability and homelessness (see Table~\ref{tab:definition_table_nonevi}). 
In total, the dataset comprises 8,000 synthetic training and development instances, alongside 616 expert-annotated test instances that encompass both synthetic and real-world clinical notes, drawn from sources such as MIMIC-IV~\cite{johnson2020mimic} and PMC-Patients~\cite{zhao2023large} (see Table~\ref{tab:data_stats}). 
This publicly available dataset not only fills a critical gap in eviction-related benchmarks but also demonstrates strong adaptability by enabling the fine-tuning of open-source LLMs for downstream use.
It offers a practical and efficient foundation for building SDoH recognition systems and integrating social care into routine clinical workflows.

\section{Results}
\label{results}

\begin{table}[!htbp]
    \vspace{-5mm}
    \centering
    \footnotesize 
    \renewcommand{\arraystretch}{1} 
    \setlength{\tabcolsep}{2pt} 
    \begin{tabularx}{\textwidth}{l|X|X|X|X}
        \toprule
        \textbf{Models} & \textbf{Synth} & \textbf{Mimic} & \textbf{PMC} & \textbf{Avg} \\
        \midrule
        \multicolumn{5}{c}{\textbf{Performance Results of Various Models for Step 1: Binary Classification (this is a balanced dataset, so we only report Micro-F1 here)}} \\
        \midrule
        GPT-4o-mini & \textcolor{blue}{0.761 (0.757-0.764) \tiny{p $<$ 0.05}} & \textcolor{blue}{0.735 (0.732-0.739) \tiny{p $<$ 0.05}} & \textcolor{blue}{0.889 (0.879-0.900) \tiny{p $<$ 0.05}} & \textcolor{blue}{0.795 (0.790-0.801) \tiny{p $<$ 0.05}} \\
        GPT-4o-mini-APO & \textcolor{blue}{0.796 (0.781-0.811) \tiny{p $<$ 0.05}} & \textcolor{blue}{0.706 (0.653-0.759) \tiny{p $<$ 0.05}} & \textcolor{blue}{0.876 (0.860-0.892) \tiny{p $<$ 0.05}} & \textcolor{blue}{0.793 (0.770-0.816) \tiny{p $<$ 0.05}} \\
        GPT-4o & \textcolor{blue}{0.880 (0.867-0.894) \tiny{p $<$ 0.05}} & \textcolor{blue}{0.860 (0.847-0.872) \tiny{p $<$ 0.05}} & \textcolor{blue}{0.839 (0.826-0.851) \tiny{p $<$ 0.05}} & \textcolor{blue}{0.859 (0.850-0.869) \tiny{p $<$ 0.05}} \\
        SynthEHR-Eviction (bert\_base\_cased) & \textcolor{red}{0.976 (0.961-0.991) \tiny{p $<$ 0.01}} & \textcolor{blue}{0.821 (0.780-0.861) \tiny{p $<$ 0.05}} & \textcolor{blue}{0.857 (0.836-0.878) \tiny{p $<$ 0.05}} & \textcolor{blue}{0.885 (0.870-0.899) \tiny{p $<$ 0.05}} \\
        SynthEHR-Eviction (biobert-v1.1)  & \textcolor{red}{\textbf{\underline{0.987 (0.979-0.995)} \tiny{p $<$ 0.01}}} & \textcolor{orange}{0.916 (0.890-0.942)} & \textcolor{blue}{0.892 (0.854-0.930) \tiny{p $<$ 0.05}} & \textcolor{orange}{0.931 (0.916-0.947)} \\
        SynthEHR-Eviction (Bio\_ClinicalBERT)  & \textcolor{red}{0.978 (0.960-0.995) \tiny{p $<$ 0.01}} & \textcolor{blue}{0.756 (0.706-0.806) \tiny{p $<$ 0.05}} & \textcolor{blue}{0.805 (0.718-0.893) \tiny{p $<$ 0.05}} & \textcolor{blue}{0.846 (0.813-0.880) \tiny{p $<$ 0.05}} \\
        SynthEHR-Eviction (LLama-3.1-8B) & \textcolor{red}{0.966 (0.956-0.976) \tiny{p $<$ 0.01}} & \textcolor{blue}{0.866 (0.850-0.882) \tiny{p $<$ 0.05}} & \textcolor{blue}{0.891 (0.875-0.907) \tiny{p $<$ 0.05}} & \textcolor{blue}{0.908 (0.897-0.919) \tiny{p $<$ 0.05}} \\
        SynthEHR-Eviction (LLama-3.2-3B) & \textcolor{orange}{0.934 (0.929-0.938)} & \textcolor{blue}{0.868 (0.858-0.877) \tiny{p $<$ 0.05}} & \textcolor{blue}{0.824 (0.813-0.834) \tiny{p $<$ 0.05}} & \textcolor{blue}{0.875 (0.871-0.879) \tiny{p $<$ 0.05}} \\
        SynthEHR-Eviction (Qwen2.5-7B) & \textcolor{red}{0.963 (0.955-0.981) \tiny{p $<$ 0.01}} & \textcolor{blue}{0.912 (0.904-0.920) \tiny{p $<$ 0.05}} & \textcolor{blue}{0.891 (0.882-0.900) \tiny{p $<$ 0.05}} & \textcolor{orange}{0.922 (0.918-0.926)} \\
        SynthEHR-Eviction (Qwen2.5-3B) & \textcolor{red}{\textbf{\underline{0.987 (0.976-0.997)} \tiny{p $<$ 0.001}}} & \textcolor{blue}{0.850 (0.840-0.859) \tiny{p $<$ 0.05}} & \textcolor{blue}{0.834 (0.820-0.848) \tiny{p $<$ 0.05}} & \textcolor{blue}{0.890 (0.884-0.896) \tiny{p $<$ 0.05}} \\
        \hline
        GPT-4o-APO (baseline) & 0.939 (0.912-0.965) & \textbf{\underline{0.938 (0.926-0.949)}} & \textbf{\underline{0.947 (0.936-0.958)}} & \textbf{\underline{0.941 (0.931-0.951)}} \\

        \midrule
    
        \multicolumn{5}{c}{\textbf{Performance Results of Various Models for Step 2: Eviction Multi-Class (Macro-F1 shown above, Micro-F1 shown below)}} \\

        \midrule

        GPT-4o-mini & \textcolor{blue}{0.337 (0.331-0.343) \tiny{p $<$ 0.05}} & \textcolor{blue}{0.238 (0.217-0.260) \tiny{p $<$ 0.05}} & \textcolor{blue}{0.243 (0.217-0.269) \tiny{p $<$ 0.05}} & \textcolor{blue}{0.273 (0.263-0.283) \tiny{p $<$ 0.05}} \\
        GPT-4o-mini-APO & \textcolor{blue}{0.760 (0.696-0.823) \tiny{p $<$ 0.05}} & \textcolor{blue}{0.672 (0.652-0.692) \tiny{p $<$ 0.05}} & \textcolor{blue}{0.641 (0.547-0.735) \tiny{p $<$ 0.05}} & \textcolor{blue}{0.691 (0.640-0.742) \tiny{p $<$ 0.05}} \\
        GPT-4o & \textcolor{blue}{0.670 (0.629-0.711) \tiny{p $<$ 0.05}} & \textcolor{blue}{0.521 (0.502-0.539) \tiny{p $<$ 0.05}} & \textcolor{blue}{0.352 (0.302-0.403) \tiny{p $<$ 0.05}} & \textcolor{blue}{0.515 (0.489-0.540) \tiny{p $<$ 0.05}} \\
        SynthEHR-Eviction (bert\_base\_cased) & \textcolor{blue}{0.806 (0.790-0.822) \tiny{p $<$ 0.05}} & \textcolor{blue}{0.784 (0.734-0.834) \tiny{p $<$ 0.05}} & \textcolor{blue}{0.384 (0.248-0.519) \tiny{p $<$ 0.05}} & \textcolor{blue}{0.658 (0.604-0.712) \tiny{p $<$ 0.05}} \\
        SynthEHR-Eviction (biobert-v1.1) & \textcolor{blue}{0.811 (0.792-0.831) \tiny{p $<$ 0.05}} & \textcolor{blue}{0.668 (0.585-0.752) \tiny{p $<$ 0.05}} & \textcolor{blue}{0.342 (0.300-0.384) \tiny{p $<$ 0.05}} & \textcolor{blue}{0.607 (0.568-0.646) \tiny{p $<$ 0.05}} \\
        SynthEHR-Eviction (Bio\_ClinicalBERT) & \textcolor{blue}{0.790 (0.762-0.818) \tiny{p $<$ 0.05}} & \textcolor{blue}{0.658 (0.630-0.686) \tiny{p $<$ 0.05}} & \textcolor{blue}{0.346 (0.321-0.371) \tiny{p $<$ 0.05}} & \textcolor{blue}{0.598 (0.584-0.611) \tiny{p $<$ 0.05}} \\
        SynthEHR-Eviction (LLama-3.1-8B) & \textcolor{orange}{0.881 (0.820-0.942)} & \textcolor{red}{0.935 (0.907-0.962) \tiny{p $<$ 0.01}} & \textcolor{orange}{\textbf{\underline{0.848 (0.821-0.876)}}} & \textcolor{orange}{\textbf{\underline{0.888 (0.872-0.904)}}} \\
        SynthEHR-Eviction (LLama-3.2-3B) & \textcolor{orange}{0.909 (0.895-0.922)} & \textcolor{orange}{0.812 (0.665-0.958)} & \textcolor{orange}{0.801 (0.774-0.828)} & \textcolor{orange}{0.840 (0.784-0.897)} \\
        SynthEHR-Eviction (Qwen2.5-7B) & \textcolor{orange}{\textbf{\underline{0.920 (0.903-0.938)}}} & \textcolor{red}{\textbf{\underline{0.939 (0.933-0.944)} \tiny{p $<$ 0.001}}} & \textcolor{orange}{0.804 (0.767-0.842)} & \textcolor{orange}{\textbf{\underline{0.888 (0.869-0.907)}}} \\
        SynthEHR-Eviction (Qwen2.5-3B) & \textcolor{orange}{0.889 (0.824-0.954)} & \textcolor{orange}{0.850 (0.827-0.873)} & \textcolor{orange}{0.825 (0.745-0.905)} & \textcolor{orange}{0.855 (0.824-0.886)} \\
        \hline
        GPT-4o-APO (baseline) & 0.918 (0.912-0.925) & 0.878 (0.846-0.911) & 0.836 (0.750-0.923) & 0.878 (0.842-0.914) \\
        
        \midrule
        GPT-4o-mini & \textcolor{blue}{0.504 (0.496-0.512) \tiny{p $<$ 0.05}} & \textcolor{blue}{0.357 (0.331-0.383) \tiny{p $<$ 0.05}} & \textcolor{blue}{0.438 (0.410-0.465) \tiny{p $<$ 0.05}} & \textcolor{blue}{0.433 (0.421-0.444) \tiny{p $<$ 0.05}} \\
        GPT-4o-mini-APO & \textcolor{blue}{0.840 (0.791-0.889) \tiny{p $<$ 0.05}} & \textcolor{blue}{0.772 (0.752-0.792) \tiny{p $<$ 0.05}} & \textcolor{blue}{0.681 (0.609-0.754) \tiny{p $<$ 0.05}} & \textcolor{blue}{0.764 (0.722-0.807) \tiny{p $<$ 0.05}} \\
        GPT-4o & \textcolor{blue}{0.754 (0.741-0.767) \tiny{p $<$ 0.05}} & \textcolor{blue}{0.583 (0.567-0.600) \tiny{p $<$ 0.05}} & \textcolor{blue}{0.502 (0.459-0.545) \tiny{p $<$ 0.05}} & \textcolor{blue}{0.613 (0.595-0.631) \tiny{p $<$ 0.05}} \\
        SynthEHR-Eviction (bert\_base\_cased) & \textcolor{blue}{0.810 (0.795-0.825) \tiny{p $<$ 0.05}} & \textcolor{orange}{0.880 (0.842-0.918)} & \textcolor{blue}{0.444 (0.285-0.603) \tiny{p $<$ 0.05}} & \textcolor{blue}{0.711 (0.647-0.775) \tiny{p $<$ 0.05}} \\ 
        SynthEHR-Eviction (biobert-v1.1) & \textcolor{blue}{0.829 (0.790-0.868) \tiny{p $<$ 0.05}} & \textcolor{blue}{0.745 (0.700-0.790) \tiny{p $<$ 0.05}} & \textcolor{blue}{0.434 (0.289-0.579) \tiny{p $<$ 0.05}} & \textcolor{blue}{0.669 (0.603-0.736) \tiny{p $<$ 0.05}} \\
        SynthEHR-Eviction (Bio\_ClinicalBERT) & \textcolor{blue}{0.791 (0.765-0.818) \tiny{p $<$ 0.05}} & \textcolor{blue}{0.775 (0.730-0.820) \tiny{p $<$ 0.05}} & \textcolor{blue}{0.390 (0.343-0.437) \tiny{p $<$ 0.05}} & \textcolor{blue}{0.652 (0.623-0.682) \tiny{p $<$ 0.05}} \\
        SynthEHR-Eviction (LLama-3.1-8B) & \textcolor{orange}{0.906 (0.888-0.923)} & \textcolor{red}{0.925 (0.909-0.941) \tiny{p $<$ 0.01}} & \textcolor{orange}{\textbf{\underline{0.861 (0.834-0.888)}}} & \textcolor{orange}{\textbf{\underline{0.897 (0.882-0.913)}}} \\
        SynthEHR-Eviction (LLama-3.2-3B) & \textcolor{orange}{0.908 (0.896-0.921)} & \textcolor{orange}{0.882 (0.848-0.916)} & \textcolor{orange}{0.824 (0.796-0.853)} & \textcolor{orange}{0.871 (0.849-0.893)} \\
        SynthEHR-Eviction (Qwen2.5-7B) & \textcolor{orange}{\textbf{\underline{0.921(0.905-0.938)}}} & \textcolor{red}{\textbf{\underline{0.938(0.932-0.945)} \tiny{p $<$ 0.001}}} & \textcolor{orange}{0.824 (0.789-0.860)} & \textcolor{orange}{0.895 (0.877-0.912)} \\
        SynthEHR-Eviction (Qwen2.5-3B) & \textcolor{orange}{0.911 (0.901-0.921)} & \textcolor{blue}{0.853 (0.832-0.875) \tiny{p $<$ 0.05}} & \textcolor{orange}{0.857 (0.801-0.914)} & \textcolor{orange}{0.874 (0.856-0.892)} \\
        \hline
        GPT-4o-APO (baseline) & 0.909(0.886-0.931) & 0.883(0.856-0.911) & 0.868(0.806-0.929) & 0.886(0.858-0.915) \\

        \midrule
        \multicolumn{5}{c}{\textbf{Performance Results of Various Models for Step 3: Non-Eviction Multi-Class (Macro-F1 shown above, Micro-F1 shown below)}} \\

        \midrule

        GPT-4o-mini & \textcolor{blue}{0.830 (0.823-0.837) \tiny{p $<$ 0.05}} & \textcolor{blue}{0.672 (0.666-0.678) \tiny{p $<$ 0.05}} & \textcolor{blue}{0.645 (0.586-0.704) \tiny{p $<$ 0.05}} & \textcolor{blue}{0.716 (0.696-0.735) \tiny{p $<$ 0.05}} \\
        GPT-4o-mini-APO & \textcolor{orange}{0.881 (0.875-0.887)} & \textcolor{blue}{0.737 (0.716-0.757) \tiny{p $<$ 0.05}} & \textcolor{blue}{0.724 (0.695-0.754) \tiny{p $<$ 0.05}} & \textcolor{blue}{0.781 (0.771-0.791) \tiny{p $<$ 0.05}} \\
        GPT-4o & \textcolor{blue}{0.850 (0.838-0.862) \tiny{p $<$ 0.05}} & \textcolor{blue}{0.795 (0.769-0.820) \tiny{p $<$ 0.05}} & \textcolor{blue}{0.746 (0.713-0.779) \tiny{p $<$ 0.05}} & \textcolor{blue}{0.797 (0.782-0.812) \tiny{p $<$ 0.05}} \\
        SynthEHR-Eviction (bert\_base\_cased) & \textcolor{blue}{0.865 (0.858-0.873) \tiny{p $<$ 0.05}} & \textcolor{blue}{0.600 (0.486-0.714) \tiny{p $<$ 0.05}} & \textcolor{blue}{0.507 (0.303-0.711) \tiny{p $<$ 0.05}} & \textcolor{blue}{0.658 (0.581-0.735) \tiny{p $<$ 0.05}} \\
        SynthEHR-Eviction (biobert-v1.1) & \textcolor{blue}{0.863 (0.851-0.874) \tiny{p $<$ 0.05}} & 0\textcolor{blue}{.623 (0.478-0.767) \tiny{p $<$ 0.05}} & \textcolor{blue}{0.564 (0.418-0.711) \tiny{p $<$ 0.05}} & \textcolor{blue}{0.683 (0.611-0.756) \tiny{p $<$ 0.05}} \\
        SynthEHR-Eviction (Bio\_ClinicalBERT) & \textcolor{blue}{0.857 (0.847-0.867) \tiny{p $<$ 0.05}} & \textcolor{blue}{0.646 (0.601-0.692) \tiny{p $<$ 0.05}} & \textcolor{blue}{0.423 (0.395-0.451) \tiny{p $<$ 0.05}} & \textcolor{blue}{0.642 (0.631-0.654) \tiny{p $<$ 0.05}} \\
        SynthEHR-Eviction (LLama-3.1-8B) & \textcolor{red}{\textbf{\underline{0.968 (0.962-0.973)} \tiny{p $<$ 0.001}}} & \textcolor{orange}{0.847 (0.821-0.874)} & \textcolor{orange}{0.833 (0.798-0.868)} & \textcolor{orange}{0.883 (0.866-0.900)} \\
        SynthEHR-Eviction (LLama-3.2-3B) & 0.914 (0.840-0.988) & \textcolor{blue}{0.642 (0.559-0.725) \tiny{p $<$ 0.05}} & \textcolor{blue}{0.782 (0.723-0.841) \tiny{p $<$ 0.05}} & 0\textcolor{blue}{.780 (0.736-0.824) \tiny{p $<$ 0.05}} \\
        SynthEHR-Eviction (Qwen2.5-7B) & \textcolor{red}{0.965 (0.958-0.973) \tiny{p $<$ 0.001}} & \textcolor{orange}{\textbf{\underline{0.864 (0.852-0.877)}}} & \textcolor{orange}{\textbf{\underline{0.879 (0.835-0.923)}}} & \textcolor{red}{\textbf{\underline{0.903 (0.888-0.918)} \tiny{p $<$ 0.05}}} \\
        SynthEHR-Eviction (Qwen2.5-3B) & \textcolor{red}{0.950 (0.942-0.958) \tiny{p $<$ 0.001}} & \textcolor{blue}{0.727 (0.646-0.808) \tiny{p $<$ 0.05}} & \textcolor{blue}{0.785 (0.749-0.821) \tiny{p $<$ 0.05}} & \textcolor{blue}{0.821 (0.787-0.855) \tiny{p $<$ 0.05}} \\
        \hline
        GPT-4o-APO (baseline) & 0.900 (0.877-0.923) & 0.850 (0.837-0.863) & 0.870 (0.827-0.913) & 0.873 (0.852-0.894) \\
                
        \midrule
        GPT-4o-mini & \textcolor{blue}{0.836 (0.830-0.842) \tiny{p $<$ 0.05}} & \textcolor{blue}{0.703 (0.698-0.708) \tiny{p $<$ 0.05}} & \textcolor{blue}{0.685 (0.669-0.701) \tiny{p $<$ 0.05}} & \textcolor{blue}{0.741 (0.734-0.748) \tiny{p $<$ 0.05}} \\
        GPT-4o-mini-APO & \textcolor{orange}{0.923 (0.904-0.942)} & \textcolor{blue}{0.783 (0.725-0.840) \tiny{p $<$ 0.05}} & \textcolor{orange}{0.852 (0.806-0.898)} & \textcolor{blue}{0.852 (0.827-0.878) \tiny{p $<$ 0.05}} \\
        GPT-4o & \textcolor{blue}{0.858 (0.851-0.866) \tiny{p $<$ 0.05}} & \textcolor{blue}{0.851 (0.839-0.864) \tiny{p $<$ 0.05}} & \textcolor{blue}{0.804 (0.791-0.817) \tiny{p $<$ 0.05}} & \textcolor{blue}{0.838 (0.831-0.844) \tiny{p $<$ 0.05}} \\ 
        SynthEHR-Eviction (bert\_base\_cased) & \textcolor{red}{0.965 (0.958-0.973) \tiny{p $<$ 0.01}} & \textcolor{blue}{0.600 (0.486-0.714) \tiny{p $<$ 0.05}} & \textcolor{blue}{0.589 (0.415-0.763) \tiny{p $<$ 0.05}} & \textcolor{blue}{0.718 (0.621-0.815) \tiny{p $<$ 0.05}} \\
        SynthEHR-Eviction (biobert-v1.1) & \textcolor{red}{0.963 (0.951-0.974) \tiny{p $<$ 0.01}} & \textcolor{blue}{0.623 (0.478-0.767) \tiny{p $<$ 0.05}} & \textcolor{blue}{0.604 (0.443-0.765) \tiny{p $<$ 0.05}} & \textcolor{blue}{0.730 (0.627-0.833) \tiny{p $<$ 0.05}} \\
        SynthEHR-Eviction (Bio\_ClinicalBERT) & \textcolor{red}{0.961 (0.956-0.966) \tiny{p $<$ 0.01}} & \textcolor{blue}{0.630 (0.612-0.648) \tiny{p $<$ 0.05}} & \textcolor{blue}{0.478 (0.459-0.497) \tiny{p $<$ 0.05}} & \textcolor{blue}{0.690 (0.683-0.696) \tiny{p $<$ 0.05}} \\
        SynthEHR-Eviction (LLama-3.1-8B) & \textcolor{red}{\textbf{\underline{0.968(0.963-0.973)} \tiny{p $<$ 0.01}}} & \textcolor{orange}{0.850(0.820-0.880)} & \textcolor{red}{\textbf{\underline{0.926 (0.910-0.942)} \tiny{p $<$ 0.05}}} & \textcolor{orange}{\textbf{\underline{0.915(0.901-0.929)}}} \\
        SynthEHR-Eviction (LLama-3.2-3B) & \textcolor{orange}{0.939 (0.917-0.960)} & \textcolor{blue}{0.817 (0.787-0.847) \tiny{p $<$ 0.05}} & \textcolor{orange}{0.874 (0.844-0.904)} & \textcolor{orange}{0.876 (0.856-0.897)} \\
        SynthEHR-Eviction (Qwen2.5-7B) & \textcolor{red}{0.965 (0.958-0.973) \tiny{p $<$ 0.01}} & \textcolor{orange}{0.866 (0.854-0.877)} & \textcolor{orange}{0.907 (0.871-0.944)} & \textcolor{red}{0.913 (0.900-0.925) \tiny{p $<$ 0.05}} \\
        SynthEHR-Eviction (Qwen2.5-3B) & \textcolor{red}{0.951 (0.944-0.959) \tiny{p $<$ 0.01}} & \textcolor{blue}{0.800 (0.769-0.831) \tiny{p $<$ 0.05}} & \textcolor{blue}{0.837 (0.818-0.856) \tiny{p $<$ 0.05}} & \textcolor{blue}{0.863 (0.849-0.877) \tiny{p $<$ 0.05}} \\
        \hline
        GPT-4o-APO (baseline) & 0.910(0.877-0.943) & \textbf{\underline{0.878(0.853-0.904)}} & 0.878(0.833-0.922) & 0.889(0.867-0.911) \\

    \bottomrule
    \end{tabularx}
    \caption{Performance comparison of various models across all steps and datasets.
    We report both Macro-F1 and Micro-F1 scores with 95\% confidence intervals for Step 1 (Binary Classification), Step 2 (Eviction Multi-Class Classification), and Step 3 (Non-Eviction Multi-Class Classification).
    \textit{SynthEHR-Eviction} refers to models fine-tuned using the SynthEHR-Eviction dataset created via our Augmenter + Annotator pipeline.
    \textit{GPT-4o-APO} serves as the baseline for statistical comparisons in each subtask.
    \textcolor{blue}{Blue cells} indicate scores that are significantly lower than the baseline,
    \textcolor{orange}{orange cells} indicate scores that are not significantly different from the baseline (i.e., at a similar level as GPT-4o-APO),
    \textcolor{red}{red cells} indicate scores that are significantly higher than the baseline,
    and \textbf{\underline{bold underlined values}} represent the best performance within each column (i.e., the highest Micro-F1 or Macro-F1 for that dataset/subtask).}
    \label{tab:main_table}
\end{table}

Our experimental results demonstrate several key findings across different classification tasks.

\subsection{Step 1: Main Results for Eviction Mention Detection via Binary Classification} 
As shown in Table~\ref{tab:main_table}, the binary classification task—distinguishing between eviction and non-eviction instances—was relatively straightforward across all model architectures.
GPT-4o, when equipped with our APO pipeline, consistently achieved the highest F1 scores across all datasets—0.939 on Synth, 0.938 on MIMIC, and 0.947 on PMC—with an overall average of 0.941.
This clearly outperformed the vanilla GPT-4o without APO (average: 0.859), highlighting the benefit of its self-prompt optimization. 
In contrast, GPT-4o-mini showed no clear gain from APO, suggesting that this smaller LLM may be less capable of benefiting from its self-reflective prompt adjustments for this task.

Open-source LLMs fine-tuned on the SynthEHR-Eviction dataset also demonstrated strong performance. Qwen2.5-7B maintained robust results across all datasets (Synth: 0.963, MIMIC: 0.912, PMC: 0.891), achieving an average F1 of 0.922. 
LLaMA-3.1-8B performed similarly well, with an average score of 0.908, and achieved particularly strong results on the Synth dataset (0.966). 
Although 3B-scale models (e.g., LLaMA-3.2-3B, Qwen2.5-3B) performed competitively on Synth, they lagged behind on PMC, indicating that smaller models may struggle with longer or more complex clinical texts.

BERT-based models fine-tuned with our SynthEHR-Eviction pipeline also yielded strong results on the Synth dataset.
Notably, bert\_base\_cased (0.976), biobert-v1.1 (0.987), and Bio\_ClinicalBERT (0.978) all outperformed GPT-4o-APO on this dataset.
While their performance was less consistent on MIMIC (ranging from 0.756 to 0.916) and generally lower than GPT-4o-APO on PMC, their overall averages remained competitive.
For instance, biobert-v1.1 achieved an average F1 of 0.931—slightly surpassing Qwen2.5-7B.
This strong performance is likely due to the relative simplicity of the binary classification task and the domain alignment of these biomedical encoders, which are well-suited for short-span eviction cue detection.

\subsection{Step 2: Main Results for Multi-Class Classification of Eviction Subcategories} 
Eviction multi-class classification is substantially more challenging than binary classification because it requires distinguishing between seven fine-grained categories of eviction-related social risk.
In this more challenging setup, two strategies demonstrated clear value: fine-tuning with the SynthEHR-Eviction dataset and DSPy-based automated
prompt optimization.

First, APO brought significant improvements to GPT-4o and GPT-4o-mini. Since the eviction task is likely not covered in the pretraining corpus of these closed-source models, this represents a classic cold-start scenario. For GPT-4o, APO lifted the average Macro-F1 from 0.515 to 0.878 and Micro-F1 from 0.613 to 0.886. Similarly, GPT-4o-mini showed large gains—Macro-F1 improved from 0.273 to 0.691, and Micro-F1 from 0.433 to 0.764. These results confirm that prompt optimization enables better task alignment and more reliable multi-class reasoning, even in the absence of model fine-tuning.

Meanwhile, models fine-tuned on SynthEHR-Eviction achieved competitive and often stronger performance than the GPT-4o-APO baseline. Qwen2.5-7B reached a Macro-F1 of 0.920 and a Micro-F1 of 0.921 on the Synth dataset, slightly surpassing GPT-4o-APO’s 0.918 and 0.909. On MIMIC, it scored 0.939 (Macro-F1) and 0.938 (Micro-F1), again marginally higher than GPT-4o-APO’s 0.878 and 0.883. On PMC, Qwen2.5-7B’s performance was more comparable, with a Macro-F1 of 0.804 vs. 0.836 and Micro-F1 of 0.824 vs. 0.868. Averaged across datasets, Qwen2.5-7B achieved 0.888 (Macro-F1) and 0.895 (Micro-F1), slightly exceeding GPT-4o-APO’s 0.878 and 0.886.
LLaMA-3.1-8B also performed robustly, with Macro-F1/Micro-F1 scores of 0.881/0.906 (Synth), 0.935/0.925 (MIMIC), and 0.848/0.861 (PMC), leading to averages of 0.888 and 0.897—again on par with or better than GPT-4o-APO. Importantly, 3B-scale models such as LLaMA-3.2-3B and Qwen2.5-3B performed competitively as well, suggesting that SynthEHR-Eviction enables scalable fine-tuning across model sizes. These 3B models achieved average Micro-F1 scores of 0.871 and 0.874, respectively, demonstrating their deployment potential in resource-constrained environments without sacrificing much performance.

In contrast, BERT-based biomedical models like BioBERT and Bio\_ClinicalBERT struggled to generalize across datasets. Their average Macro-F1 and Micro-F1 scores remained below 0.70 and 0.72, respectively, with especially poor results on PMC. This suggests that while these models may encode biomedical terminology, they lack the representational flexibility needed for nuanced, fine-grained classification tasks like eviction status detection.

\subsection{Step 3: Main Results for Multi-Class Classification of Non-Eviction Subcategories} 
Unlike the eviction-specific classification in Step 2, the Step 3 task encompasses a broader range of social risk categories, including homelessness, housing instability, food insecurity, transportation insecurity, and others. 
Many of these categories are commonly documented in public health and clinical records, making them more familiar to general-purpose LLMs.
This is reflected in the relatively strong performance of GPT-4o even without APO, which achieved average scores of 0.797 Macro-F1 and 0.838 Micro-F1—significantly higher than its performance in Step 2.

Nonetheless, prompt optimization continued to enhance closed-source models. 
GPT-4o-APO reached 0.873 Macro-F1 and 0.889 Micro-F1, improving over the non-optimized version. 
The GPT-4o-mini also benefited from APO, with Macro-F1 increasing from 0.716 to 0.781, and Micro-F1 from 0.741 to 0.852.

Fine-tuned models based on SynthEHR-Eviction also demonstrated excellent performance.
Most notably, Qwen2.5-7B achieved a Macro-F1 of 0.903 and a Micro-F1 of 0.913, both significantly outperforming the GPT-4o-APO baseline (p < 0.05).
LLaMA-3.1-8B matched this strong performance with 0.883 and 0.915, respectively. 
These results highlight that even when tasks are not strictly “cold-start,” fine-tuning with task-specific high-quality data remains valuable for enhancing performance.

Encouragingly, 3B-scale models also performed well. Qwen2.5-3B achieved 0.821 (Macro-F1) and 0.863 (Micro-F1), while LLaMA-3.2-3B followed with 0.780 and 0.876, respectively—both clearly surpassing GPT-4o-mini-APO. These results demonstrate that SynthEHR-Eviction facilitates efficient scaling, enabling smaller models to reach production-quality performance without reliance on massive human annotation resources.

Finally, the biomedical BERT variants again underperformed across metrics (Macro-F1 < 0.69, Micro-F1 < 0.74), with particularly weak results on MIMIC and PMC. 
These models may lack the generative flexibility and contextual reasoning needed for multi-class prediction across heterogeneous SDoH categories.

In summary, while Step 3 categories are more likely to have been seen during LLM pretraining, SynthEHR-Eviction still offers substantial fine-tuning benefits. 
When combined with its superior performance on unseen (eviction-specific) categories, this suggests that our solution serves as a robust and broadly applicable resource for improving SDoH modeling in both known and novel scenarios.

\begin{table}[!ht]
    \centering
    \footnotesize 
    \renewcommand{\arraystretch}{1} 
    \setlength{\tabcolsep}{2pt} 
    \scalebox{1}{
    \begin{tabularx}{\textwidth}{l|l|X|X|X|X|X|X|X|X}
        \toprule
        & & \multicolumn{2}{c|}{\textbf{Synth}} & \multicolumn{2}{c|}{\textbf{Mimic}} & \multicolumn{2}{c|}{\textbf{PMC}} & \multicolumn{2}{c}{\textbf{Avg}} \\
        \midrule
        \textbf{Model} & \textbf{Class} & \textbf{MCC} & \textbf{F1} & \textbf{MCC} & \textbf{F1} & \textbf{MCC} & \textbf{F1} & \textbf{MCC} & \textbf{F1} \\
        \midrule
        GPT-4o-APO & Eviction\_absent & 1.00 & 1.00 & & & & & 1.00 & 1.00 \\
        & Eviction\_hypothetical & 1.00 & 1.00 & 0.75 & 0.75 & 0.83 & 0.84 & 0.86 & 0.86 \\
        & Eviction\_mr\_current & 0.82 & 0.84 & 0.89 & 0.90 & 0.95 & 0.96 & 0.89 & 0.90 \\
        & Eviction\_mr\_history & 0.81 & 0.81 & 0.88 & 0.89 & 0.89 & 0.89 & 0.86 & 0.86 \\
        & Eviction\_pending & 0.95 & 0.96 & 0.81 & 0.83 & 0.81 & 0.83 & 0.86 & 0.87 \\
        & Eviction\_present\_current & 0.91 & 0.93 & 0.97 & 0.97 & 0.88 & 0.89 & 0.92 & 0.93 \\
        & Eviction\_present\_history & 0.94 & 0.94 & 0.97 & 0.98 & 0.95 & 0.96 & 0.95 & 0.96 \\
        & Overall\_(macro\_f1 / micro\_f1) & 0.92 & 0.93 / 0.93 & 0.88 & 0.89 / 0.89 & 0.88 & 0.89 / 0.90 & 0.89 & 0.90 / 0.91 \\	
        \midrule

        Llama-3.1-8B & Eviction\_absent & 1.00 & 1.00 & & & & & 1.00 & 1.00 \\
        & Eviction\_hypothetical & 0.97 & 0.98 & 0.94 & 0.95 & 0.91 & 0.92 & 0.94 & 0.95 \\
        & Eviction\_mr\_current & 0.77 & 0.80 & 0.94 & 0.95 & 0.89 & 0.90 & 0.87 & 0.88 \\
        & Eviction\_mr\_history & 0.87 & 0.89 & 0.94 & 0.95 & 0.76 & 0.80 & 0.86 & 0.88 \\
        & Eviction\_pending & 0.87 & 0.88 & 0.94 & 0.95 & 0.81 & 0.83 & 0.87 & 0.89 \\
        & Eviction\_present\_current & 0.88 & 0.90 & 1.00 & 1.00 & 0.74 & 0.76 & 0.88 & 0.89 \\
        & Eviction\_present\_history & 0.91 & 0.92 & 1.00 & 1.00 & 0.85 & 0.87 & 0.92 & 0.93 \\
        & Overall\_(macro\_f1 / micro\_f1) & 0.89 & 0.80 / 0.91 & 0.96 & 0.97 / 0.97 & 0.82 & 0.85 / 0.85 & 0.89 & 0.87 / 0.91\\	
        \midrule
        Llama-3.2-3B & Eviction\_absent & 1.00 & 1.00 & & & & & 1.00 & 1.00 \\
        & Eviction\_hypothetical & 1.00 & 1.00 & 0.94 & 0.95 & 0.94 & 0.95 & 0.96 & 0.97 \\
        & Eviction\_mr\_current & 0.81 & 0.83 & 0.90 & 0.92 & 0.88 & 0.89 & 0.86 & 0.88 \\
        & Eviction\_mr\_history & 0.87 & 0.89 & 0.92 & 0.93 & 0.79 & 0.82 & 0.86 & 0.88 \\
        & Eviction\_pending & 0.89 & 0.91 & 0.92 & 0.93 & 0.83 & 0.85 & 0.88 & 0.90 \\
        & Eviction\_present\_current & 0.89 & 0.89 & 0.80 & 0.81 & 0.75 & 0.76 & 0.81 & 0.82 \\
        & Eviction\_present\_history & 0.91 & 0.92 & 0.92 & 0.93 & 0.89 & 0.91 & 0.91 & 0.92 \\
        & Overall (macro\_f1 / micro\_f1) & 0.91 & 0.92 / 0.92 & 0.90 & 0.91 / 0.92 & 0.84 & 0.86 / 0.87 & 0.89 & 0.90 / 0.90\\
        \midrule
        Qwen2.5-7B & Eviction\_absent & 1.00 & 1.00 & & & & & 1.00 & 1.00 \\
        & Eviction\_hypothetical & 0.97 & 0.98 & 0.91 & 0.92 & 0.88 & 0.89 & 0.92 & 0.93 \\
        & Eviction\_mr\_current & 0.87 & 0.88 & 0.94 & 0.95 & 0.94 & 0.95 & 0.91 & 0.93 \\
        & Eviction\_mr\_history & 0.94 & 0.94 & 0.94 & 0.95 & 0.85 & 0.87 & 0.91 & 0.92 \\
        & Eviction\_pending & 0.89 & 0.91 & 0.87 & 0.89 & 0.77 & 0.80 & 0.85 & 0.87 \\
        & Eviction\_present\_current & 0.91 & 0.93 & 0.94 & 0.94 & 0.71 & 0.73 & 0.85 & 0.87 \\
        & Eviction\_present\_history & 0.91 & 0.92 & 1.00 & 1.00 & 0.88 & 0.90 & 0.93 & 0.94 \\
        & Overall (macro\_f1 / micro\_f1) & 0.93 & 0.94 / 0.94 & 0.93 & 0.94 / 0.94 & 0.84 & 0.86 / 0.86 & 0.90 & 0.91 / 0.91 \\
        \midrule
        Qwen2.5-3B & Eviction\_absent & 1.00 & 1.00 & & & & & 1.00 & 1.00 \\
        & Eviction\_hypothetical & 0.91 & 0.93 & 0.81 & 0.82 & 0.88 & 0.89 & 0.87 & 0.88 \\
        & Eviction\_mr\_current & 0.84 & 0.86 & 0.87 & 0.89 & 0.92 & 0.93 & 0.88 & 0.89 \\
        & Eviction\_mr\_history & 0.87 & 0.88 & 0.88 & 0.90 & 0.85 & 0.86 & 0.87 & 0.88 \\
        & Eviction\_pending & 0.87 & 0.89 & 0.79 & 0.82 & 0.77 & 0.80 & 0.81 & 0.84 \\
        & Eviction\_present\_current & 0.91 & 0.93 & 0.84 & 0.85 & 0.78 & 0.79 & 0.84 & 0.85 \\
        & Eviction\_present\_history & 0.91 & 0.92 & 0.95 & 0.95 & 0.92 & 0.93 & 0.92 & 0.93 \\
        & Overall (macro\_f1 / micro\_f1) & 0.90 & 0.91 / 0.91& 0.86 & 0.87 / 0.88 & 0.85 & 0.87 / 0.87 & 0.87 & 0.88 /0.89 \\					
    \bottomrule
    \end{tabularx}
    }
    \caption{Model performance comparison across different datasets (Synth, Mimic, PMC) with MCC and F1 scores (including Macro-F1 and Micro-F1 in Overall) for each eviction class. 
    Results present four models: GPT-4o-APO, Llama-3.1-8B-FT, Llama-3.2-3B-FT, Qwen2.5-7B-FT, and Qwen2.5-3B-FT, with overall average metrics. 
    Note that "Eviction\_absent" appears exclusively in the Synth dataset as it functions as a critical negative class that establishes decision boundaries between eviction-relevant and irrelevant content during model training. Mimic and PMC datasets deliberately omit this class to focus evaluation specifically on discrimination capabilities between different positive eviction scenarios.
    Importantly, this omission is not due to data scarcity. 
    On the contrary, most real-world clinical notes in MIMIC and PMC would fall under the "Eviction\_absent" category. 
    However, including such easy negatives would inflate overall accuracy and obscure the model's ability to distinguish among nuanced positive classes, so they were excluded by design.
    }
    \label{tab:class_variation_table}
\end{table}

\subsection{Class Performance Variations Across Eviction Categories}

Table~\ref{tab:class_variation_table} reports the class-wise performance of five models—GPT-4o-APO, Llama-3.1-8B, Llama-3.2-3B, Qwen2.5-7B, and Qwen2.5-3B—on seven nuanced eviction-related categories across three datasets (Synth, Mimic, PMC). 
These classes vary in eviction status (e.g., \texttt{present}, \texttt{pending}, \texttt{hypothetical}), legal framing (e.g., \texttt{mutual rescission}), and temporal cues (e.g., \texttt{current} vs. \texttt{history}).

All models achieved perfect classification (F1 = 1.00) on \textbf{Eviction\_absent} in the Synthetic dataset. This class was excluded from Mimic and PMC to prevent dominance of trivial negatives and to emphasize the model's ability to differentiate among positive cases.

\textbf{Eviction\_hypothetical}, representing projected or threatened eviction, posed moderate challenges.
GPT-4o-APO reached an average F1 of 0.86, while several fine-tuned open-source LLMs—specifically Qwen2.5-7B and Llama-3.2-3B—surpassed this benchmark with Synth scores of 0.98 and 1.00, and solid generalization in Mimic (0.92 and 0.95), highlighting the benefit of domain-specific fine-tuning on SynthEHR-Eviction.

\textbf{Eviction\_mr\_current} and \textbf{Eviction\_mr\_history}, which capture mutually agreed lease terminations, were consistently well-handled. Qwen2.5-7B-FT achieved 0.93 and 0.92 on Macro-F1 and Micro-F1, respectively, slightly outperforming GPT-4o-APO (F1 = 0.90 and 0.86). Notably, fine-tuned 3B models also maintained high accuracy in these categories (F1 is about 0.88), suggesting strong transfer of structured legal signals even in smaller models after fine-tuning.

\textbf{Eviction\_pending}, indicating formal but unresolved eviction processes, remained somewhat ambiguous. Qwen2.5-7B and Llama-3.1-8B achieved F1 = 0.91 and 0.95 in Synth and Mimic respectively, surpassing GPT-4o-APO's average of 0.87. However, smaller models showed more variance across datasets, especially in PMC.

\textbf{Eviction\_present\_current} and \textbf{Eviction\_present\_history}, which rely on subtle differences in recency and documentation of completed evictions, remained the most challenging. 
In PMC, for example, Llama-3.1-8B dropped to 0.76 on "present\_current", while GPT-4o-APO maintained a stronger F1 of 0.93. 
These results highlight that while fine-tuned LLMs can excel in categories grounded in legal constructs, temporal ambiguity still presents difficulties.

In summary, class-level comparisons reveal that fine-tuned open-source LLMs—especially Qwen2.5-7B and Llama-3.2-3B—can match or exceed GPT-4o-APO in multiple eviction categories after training on SynthEHR-Eviction. 
Larger models (7B/8B) showed better generalization across datasets, 
while smaller models (3B) still offered strong performance on well-defined categories. 
GPT-4o-APO provided a competitive baseline with consistent performance across the board, though not uniformly superior to well-trained open models.

\begin{figure}[!htbp]
    \centering
    \includegraphics[width=0.4\linewidth]{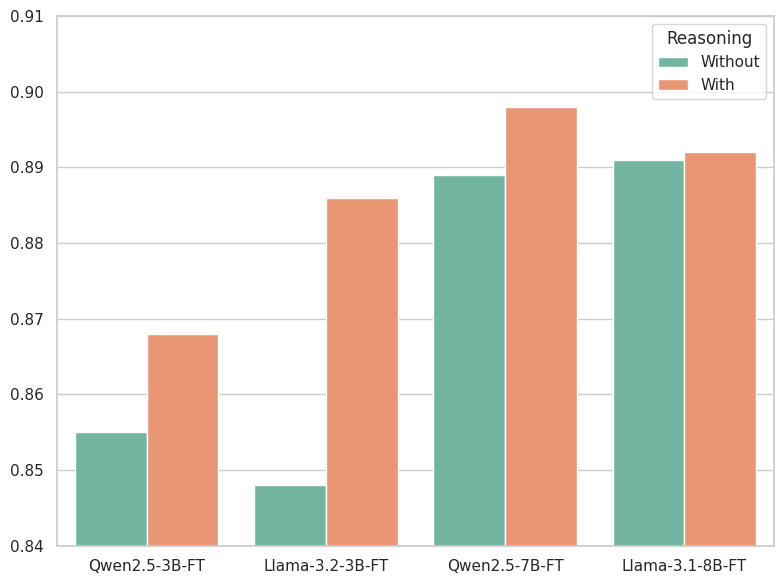}
    \caption{Performance comparison of LLMs trained with and without reasoning annotations on the Eviction Multi-Class task.}
    \label{fig:reasoning_impact_table}
\end{figure}

\subsection{Impact of Reasoning in Training Data}

To evaluate the effect of explicit reasoning annotations in training data, we prepared two versions of each model’s training set: one with reasoning explanations and one without, keeping the labels identical.
As shown in Figure~\ref{fig:reasoning_impact_table}, smaller models benefit more noticeably from reasoning annotations. For instance, Llama-3.2-3B-FT improved from 0.848 to 0.886 (+3.8), and Qwen2.5-3B-FT improved from 0.855 to 0.868 (+1.3). In contrast, larger models like Qwen2.5-7B-FT and Llama-3.1-8B-FT maintained stable performance with or without reasoning, with minimal differences (less than 0.01).
These findings suggest that the reasoning annotations generated through the SynthEHR-Eviction pipeline can provide varying degrees of structured guidance, especially for smaller models that benefit from clearer decision logic in complex classification tasks. Additionally, models trained with these annotations are capable of generating their own reasoning during inference, improving transparency and interpretability. This makes them more amenable to expert review and downstream validation, moving beyond black-box eviciton prediction.

\begin{figure}[!htbp]
    \centering
    \includegraphics[width=1\linewidth]{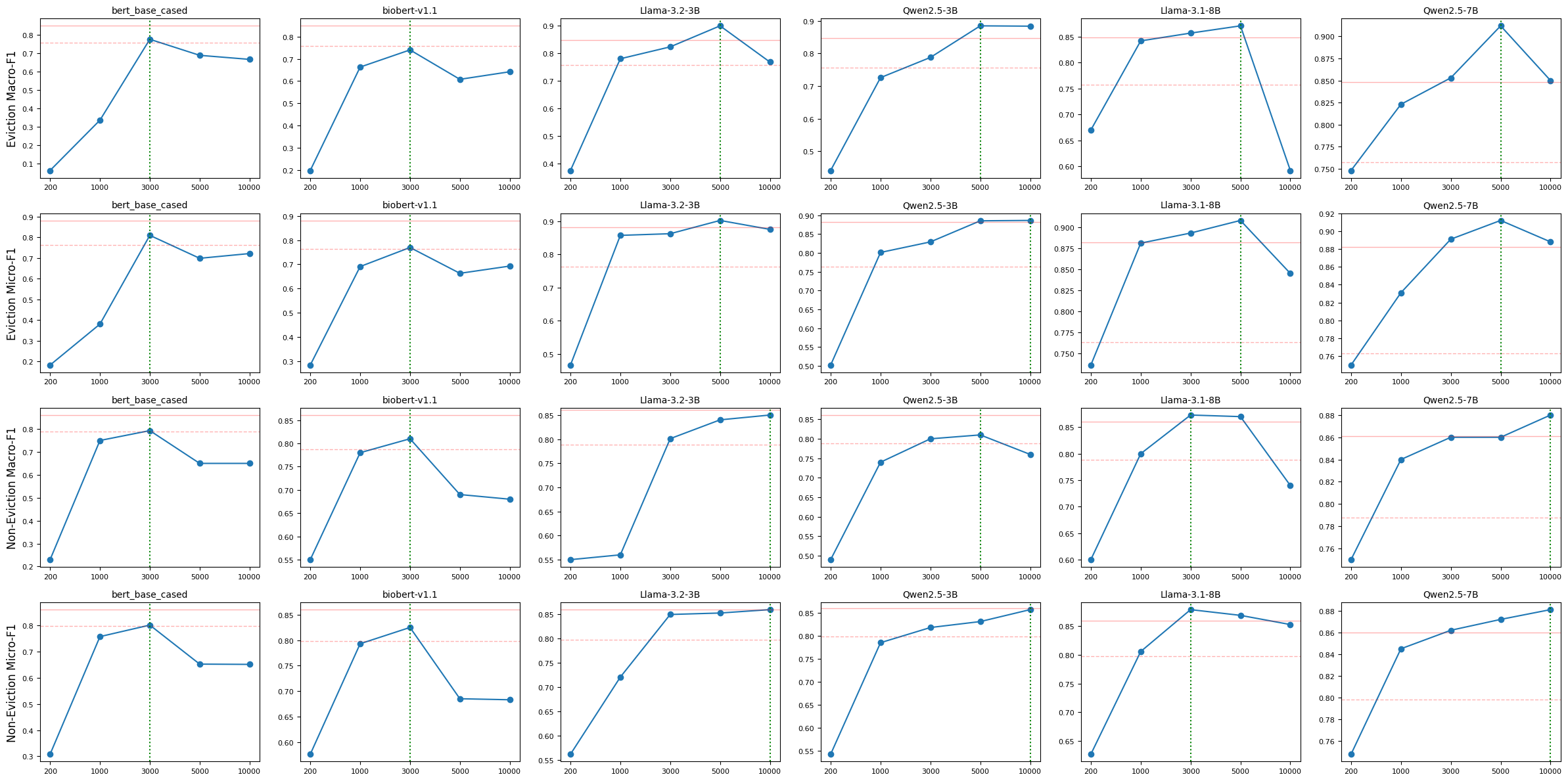}
    \caption{Impact of training set size on model performance across eviction and non-eviction classification tasks.
    Each subplot illustrates how model performance varies with different training set sizes (x-axis) across four classification scenarios: eviction (top two rows) and non-eviction (bottom two rows), evaluated using two metrics—Macro-F1 (first and third rows) and Micro-F1 (second and fourth rows). The y-axis represents the corresponding F1 score.
    Green vertical dashed lines indicate the training data size at which each model achieves its highest performance.
    Red solid and Red dashed horizontal lines denote the performance of GPT-4o-APO and GPT-4o-mini-APO, serving as a reference here.}
    \label{fig:trainset_size_overall}
\end{figure}


    

    

\subsection{Training Data Size Impact on Finetuning}

To examine how training data size affects model performance, we conducted systematic experiments on two traditional BERT-based models (BERT and BioBERT) and four open-source LLMs (two 3B-scale and two 7B/8B-scale LLMs).
We evaluated both the Eviction Multi-Class and Non-Eviction Multi-Class tasks using training datasets of varying sizes (200 to 10,000 samples) and reported both Macro-F1 and Micro-F1 scores (Figure~\ref{fig:trainset_size_overall}).

Across both tasks, models demonstrated substantial gains when the training data was increased from 200 to 3,000 samples. 
For instance, Qwen2.5-7B improved its Macro-F1 on the Eviction task from 0.748 to 0.853, and on the Non-Eviction task from 0.75 to 0.891. 
Similarly, LLaMA-3.1-8B increased from 0.670 to 0.857 (Eviction) and from 0.736 to 0.893 (Non-Eviction).
Performance gains typically plateaued beyond 5,000 samples, and several models exhibited signs of overfitting at 10,000 samples. For example, Qwen2.5-7B reached its peak Macro-F1 of 0.912 on the Eviction task at 5,000 samples, then dropped to 0.850 at 10,000 samples.

Optimal dataset sizes varied by task. For Step 2 (Eviction), LLMs such as Qwen2.5-7B required larger datasets to reach optimal performance (e.g., 0.912 Macro-F1 at 5,000 samples). In Step 3 (Non-Eviction), most models converged earlier, reaching their peak performance around 3,000 samples (e.g., Qwen2.5-7B at 0.891).

For practical fine-tuning, training with 3,000–5,000 samples provides a strong balance between performance and efficiency, since several LLMs trained on just 3,000–5,000 samples reached performance levels comparable to GPT-4o-APO.
Accordingly, we used 5,000 samples for LLMs and 3,000 samples for traditional BERT models in Step 2. For Step 3, we used 3,000 samples for both model types, as performance had largely converged at this size.

\section{Discussion}
\label{discussion}

Eviction is a critical yet underrepresented SDoH, associated with cascading adverse outcomes such as homelessness, chronic stress, and emergency care overuse~\cite{Desmond2015,Desmond2016,Desmond2017,Tsai2019,tsai2024eviction,yao2023eviction}.
While prior NLP efforts have focused on broader housing-related issues, such as homelessness or housing instability~\cite{guevara2024large,patra2021sdohextraction}, the status of eviction has been largely neglected in public clinical datasets and benchmarks.
Only very recently has this issue received direct attention—for instance, Yao et al. (2023) introduced an eviction classifier trained on a private Veterans Affairs dataset~\cite{yao2023eviction}.
To help fill this gap, we present SynthEHR-Eviction: a scalable and clinically grounded data generation pipeline that produces high-fidelity, eviction-labeled clinical notes aligned with real-world documentation practices. 
The pipeline combines LLM-based data augmentation, automated prompt optimization, and human-in-the-loop validation to synthesize instruction-style notes paired with labels derived from ICD-10 Z59 codes (e.g., Z59.0–Z59.9, covering housing and economic instability).
This label schema enhances semantic precision and interoperability with structured health data systems~\cite{cms_sdoz2023,guevara2024large}. 
Furthermore, our taxonomy goes beyond binary classification by capturing nuanced eviction subtypes—such as pending status, mutual rescission, and temporal distinctions (e.g., current vs. history)—allowing for more accurate model training and more actionable risk detection. 
By releasing both the annotated dataset and labeling guidelines, we support the development and evaluation of eviction-aware NLP tools that generalize across institutions and reflect realistic documentation variability.

One of the most practical contributions of SynthEHR-Eviction lies in its efficiency: it dramatically reduces the human effort required to construct large-scale, high-quality annotated datasets.
Manual annotation of complex SDoH categories—particularly those like eviction that involve temporal ambiguity, legal framing, and implicit cues—is notoriously labor-intensive. In our controlled study, full manual rewriting of 8,000 EHR notes took over 266 hours, and even direct annotation of labels consumed 33 hours.
In contrast, our GPT-assisted, human-in-the-loop workflow—with targeted prompt engineering, batch validation, and automated reasoning trace generation—achieved comparable data quality with less than 6 hours of human effort, representing an over 80\% reduction in annotation time. 
These findings align with recent work showing the scalability and reliability of weak supervision, prompt-based augmentation, and LLM-powered data generation in other domains~\cite{hsu2025leveraging,zhang2022survey,zha2025data,wang2022promda}.
More importantly, our pipeline offers structured reasoning alongside each label, allowing smaller models to learn decision rationales explicitly—an advantage not typically found in weakly labeled or distant supervision setups. 
Because the pipeline is built on modular components (e.g., task-adaptive prompt optimization, reasoning-based CoT supervision, and human verification loops), it generalizes to other low-resource SDoH domains where fine-grained annotation is scarce but clinically significant (e.g., utility shutoff, food insecurity, or intimate partner violence). 
As large-scale clinical NLP increasingly shifts toward low-cost, explainable, and domain-adaptable solutions, SynthEHR-Eviction provides a blueprint for scaling up dataset construction without compromising semantic depth or real-world alignment~\cite{guevara2024large,magister2022teaching}.

To systematically evaluate model generalization in eviction-related SDoH classification, we tested performance across three distinct clinical datasets—Synth, MIMIC, and PMC—that vary substantially in linguistic style, narrative density, and documentation patterns (Table~\ref{tab:note_sample_table}).
Our results reveal a consistent domain shift: models achieved their highest performance on the synthetic data (Synth), performed moderately on real-world EHRs (MIMIC), and worst on academic case reports (PMC). While all three datasets adopt instruction-style formatting, they differ in vocabulary, structure, and contextual framing of eviction signals. 
The Synth dataset, generated using our LLM-based pipeline, closely mirrors the model’s generation style and served as the primary training source, explaining its strong in-domain performance.
In contrast, MIMIC notes—written by clinicians—feature more heterogeneous phrasing (e.g., “forced to remove from the rented house”) and exhibit subtle lexical shifts. 
The PMC notes pose the greatest challenge, as eviction information is often deeply embedded in long diagnostic narratives and may only appear incidentally in unrelated sections, requiring fine-grained relevance detection and temporal disambiguation (As shown in Table~\ref{tab:note_sample_table}).
This distributional mismatch explains the observed drop in generalization: for example, Qwen2.5-7B’s Macro-F1 fell from 0.920 (Synth) to 0.804 (PMC). 
These findings are consistent with prior SDoH extraction studies documenting performance degradation when models are evaluated on free-text narratives~\cite{keloth2025social,han2022sdohclassification,mitra2023sdohsuicide}.
Notably, a simple data composition intervention—replacing just 30\% of synthetic training data with real PMC notes—led to substantial gains in out-of-domain performance. On the PMC portion of the Step 2 task, Qwen2.5-7B’s Micro-F1 improved by +0.302 (Table~\ref{tab:training_composition_table}).
This result reinforces a growing consensus in clinical NLP: while synthetic data offers scale and control, it is insufficient on its own for achieving robust generalization. Incorporating even a small proportion of real-world examples into training is essential for building models that transfer reliably to naturalistic clinical settings~\cite{guevara2024large}.
Moreover, because SynthEHR-Eviction spans synthetic and real-world document styles, it enabled us to evaluate how different model families generalize across varied clinical documentation settings.
On the challenging PMC multi-class classification task, a general-domain LLaMA-3.1-8B model achieved a Micro-F1 of 0.861, while the BERT-base-cased model fell to just 0.444—a difference of over 40 percentage points.
This substantial gap underscores LLMs’ superior ability to handle diverse and noisy clinical narratives, likely due to their larger parameter counts, broader pretraining corpora, and emergent reasoning capabilities~\cite{guevara2024large}.
These findings are in line with recent benchmarks showing that state-of-the-art LLMs, such as GPT-4 and Flan-T5-XXL, consistently outperform domain-specific biomedical models like BioBERT and ClinicalBERT across SDoH extraction and other fine-grained clinical NLP tasks~\cite{majid2025evaluating,guevara2024large}. 
However, we also recognize that LLM deployment in real-world settings presents nontrivial challenges: inference latency, compute costs, and limited GPU infrastructure may hinder adoption in under-resourced health systems~\cite{dennstadt2025implementing}. 
Thus, future work should explore hybrid strategies such as model distillation, dynamic inference control, or task-specific adapter modules to achieve a more practical balance between performance and feasibility.
A nuanced understanding of this tradeoff is critical to ensuring that SDoH-aware NLP tools are not only accurate but also equitable and deployable at scale.

Our fine-grained eviction label schema—distinguishing between categories such as \texttt{Eviction\_present\_current} and \texttt{Eviction\_present\_history}—allowed us to examine one of the most persistent challenges in clinical NLP: temporal reasoning errors arising from implicit event timelines in free-text notes. 
Temporal ambiguity emerged as a consistent source of classification error during model evaluation.
A closer inspection of eviction status classification errors (Cases 1 and 2, Table~\ref{tab:case_study_table}) reveals a critical challenge: the models’ difficulty in distinguishing between historical and current eviction events due to implicit temporal ambiguities in clinical notes.
In Case 1, the phrase "resides now" prompted the Llama-3.1-8B model to incorrectly classify a historical eviction (\texttt{Eviction\_present\_history}) as ongoing (\texttt{Eviction\_present\_current}). 
Conversely, in Case 2, the model incorrectly labeled a recent eviction ("two months ago") as historical, heavily influenced by the initial lexical cue "history," despite the explicit recent temporal context.
These bidirectional error patterns emphasize temporal ambiguity, where historical events can have ongoing impacts or be conflated with current circumstances. 
Notably, our observations align with known challenges in clinical NLP: reasoning about the timing of events in narratives is difficult for models to accurately capture~\cite {kougia2024analysing}. 
Even advanced LLMs have been reported to struggle with temporal relation extraction, performing worse than fine-tuned models and showing inconsistent handling of event order. 
This persistent ambiguity has been documented since earlier efforts, such as THYME~\cite{styler2014temporal} and the Clinical TempEval challenges~\cite{bethard-etal-2016-semeval}, and remains a focus of current research.
Our findings reinforce that temporal context (e.g., distinguishing a resolved eviction in the past from a patient’s current housing crisis) is a critical source of error. 
Future improvements may require incorporating explicit temporal tagging or leveraging note timestamps—potentially guided by recent time-aware training frameworks~\cite{liu2025time}—to help models disambiguate event timelines.
Addressing these temporal reasoning issues is vital, as accurately determining whether an eviction is ongoing or historical is crucial for effective risk assessment and intervention.

We also explored the use of structured reasoning annotations—i.e., chain-of-thought explanations provided during training—to examine their utility in enhancing model performance and interpretability, particularly for smaller, resource-efficient models. 
Our experiments showed that smaller models benefit disproportionately from this form of supervision: LLaMA-3.2-3B improved by +3.8 Macro-F1 and Qwen2.5-3B by +1.3, compared to their counterparts trained on labels alone.
In contrast, larger models (Llama-3.1-8B, Qwen2.5-7B) showed minimal improvements under the same setup, indicating that extensive pretraining had already endowed them with strong inherent reasoning abilities.
These findings are consistent with prior observations on chain-of-thought prompting~\cite{wei2022chain}: smaller models often struggle to exhibit complex reasoning capabilities unless explicitly guided, while larger models tend to benefit less from such supervision. 
SynthEHR-Eviction operationalizes this insight by pairing each label with interpretable reasoning traces, allowing smaller models to learn not just what the correct answer is, but why it is correct.
Our results reinforce this trend, demonstrating that reasoning-guided training can substantially improve the performance of smaller models, while offering limited additional value for larger ones. 
This performance disparity helps explain why our 3B-scale models required explicit reasoning supervision to accurately distinguish fine-grained eviction categories, while the 7–8B models performed comparably well without it.
Qualitative analysis further supports this conclusion. In Case 3 (Table~\ref{tab:case_study_table}), the Llama-3.2-3B model trained with reasoning annotations correctly classified a complex eviction situation as \texttt{Eviction\_pending}, clearly articulating the active and unresolved nature of the eviction process. 
In contrast, the same model trained without reasoning annotations misclassified the scenario as \texttt{Eviction\_present\_current}, failing to capture the subtle temporal and procedural cues.
This example illustrates how structured reasoning guidance can help smaller models develop more precise and interpretable decision boundaries in semantically challenging cases.
From a deployment perspective, these results highlight the practical value of reasoning annotations in improving both the performance and interpretability of smaller, more resource-efficient models~\cite{mitra2023orca}.
By leveraging the reasoning-enhanced supervision provided in SynthEHR-Eviction, smaller models can match or exceed the performance of much larger LLMs, while remaining compatible with edge deployments and low-resource clinical infrastructures.
This is particularly relevant in real-world clinical environments where compute budgets are constrained, such as community hospitals, mobile health units, or edge devices operating in under-resourced settings~\cite{dennstadt2025implementing}. 
Reasoning-augmented training, as implemented in our pipeline, thus enables scalable and equitable NLP solutions for SDoH extraction.

While structured reasoning annotations improved model transparency, our evaluation further revealed limitations in reasoning fidelity—that is, whether a model’s explanation faithfully reflects its underlying decision process.
Our case studies (Cases 4–6, Table~\ref{tab:case_study_table_2}) revealed interpretability challenges even when the final classification was correct.
In several instances, the model produced an explanation that was partially incorrect or conceptually misaligned, despite outputting the right label.  For example, the Llama-3.1-8B model correctly labeled a case as \texttt{Eviction\_mr\_current} (mutual rescission current), but its explanation erroneously stated that the mutual lease rescission was unrelated to eviction, indicating a misunderstanding of the concept (Case 4). 
In another case (Case 5), the model correctly identified the eviction as \texttt{Eviction\_pending} but described it using language that suggests a historical context (“so we will label it as ‘history’”), which deviates from the annotation guidelines.
Similarly, a note properly labeled as \texttt{Eviction\_present\_current} was explained as if the eviction process were still ongoing, failing to recognize it had been completed. 
These examples demonstrate that the presence of a reasoning chain does not guarantee its fidelity—that is, alignment with the model’s actual decision boundary or reasoning strategy. 
Recent studies~\cite{yeo2024interpretable,bilal2025llms} on LLM explainability have similarly noted that models often generate plausible-sounding rationales that are not fully faithful to their internal computations, or even inconsistent with their final outputs. 
In other words, an AI can be “right for the wrong reasons,”~\cite{agarwal2024faithfulness,jin2024hidden,yang2025unveiling} as seen in our case analyses. 
This reinforces the need for rigorous human oversight and domain-informed evaluation when reasoning annotations are used for interpretability. 
SynthEHR-Eviction provides such a platform, where both correctness and explanation quality can be jointly assessed.
Our findings echo growing efforts to measure and improve the faithfulness of chain-of-thought explanations in LLMs~\cite{paul2024making}. 
Ensuring that the model’s explanations are not only present but also correct and useful is an important direction for future work, especially in high-stakes clinical applications~\cite{abgrall2024should}. 
This aligns with broader efforts in clinical NLP to reduce hallucinations and improve factual alignment in model-generated outputs~\cite{kim2025medical,ji2023survey,mishra2024synfac,yao2023improving}.
Continual refinement of the reasoning templates and collaboration with clinical experts will be needed to close these gaps~\cite{yang2025unveiling,jin2024hidden}.

Finally, our experiments suggest that model performance tends to level off between 3,000 and 5,000 training examples, with little improvement—and in some cases, slight overfitting—when more data is added.
This threshold may reflect either the high informational density of our pipeline-generated data, where critical semantic cues are well-covered early, or diminishing marginal returns from increasingly repetitive synthetic notes. 
These hypotheses warrant further investigation. Nonetheless, our results suggest that 3,000–5,000 examples may offer a practical trade-off between data curation effort and fine-tuning performance for this task.

While our results demonstrate the promise of SynthEHR-Eviction, several limitations remain.
First, the synthetic training data, although grounded in real EHR patterns, may not capture the full variability and sociolinguistic nuance found in actual clinical documentation across all institutions and populations. This raises potential concerns about distributional bias and overfitting to synthetic patterns.
Second, we did not exhaustively explore all methods for generating synthetic data. We focused on prompt optimization and high-precision augmentation strategies that are reproducible and low-cost, which may exclude higher-recall alternatives.
Third, although the initial results are promising, additional clinical validation will be necessary to establish the model's performance in real-world environments.
Finally, as with any annotation effort, the quality of the output is contingent upon the guidelines and decisions of human annotators. Even with structured reasoning supervision, SDoH annotation remains a complex task subject to ambiguity and inter-annotator variability.
Future work should investigate how to balance synthetic and real data, address biases introduced by LLM pretraining, and assess the downstream utility of eviction detection in improving health outcomes and promoting equity.

Taken together, SynthEHR-Eviction provides a scalable and clinically grounded foundation for enhancing the detection of eviction-related SDoH in unstructured clinical notes.
With its structured reasoning supervision, domain-diverse testbed, and interpretable outputs, the benchmark is particularly well-suited for evaluating and refining NLP models in socially sensitive, high-stakes settings. 
Beyond its methodological contributions, SynthEHR-Eviction introduces two broadly applicable design principles—domain-specific abstraction and task-aligned supervision—that can be extended to other underrepresented SDoH domains, such as legal hardship or caregiving burden.
Its automated and modular pipeline supports low-cost construction of semantically rich datasets, enabling scalable integration of diverse social contexts into downstream clinical NLP applications.
For example, when applied to AI hospital simulations, these SDoH-aware datasets can enrich the behavioral profiles of patient agents~\cite{yao2025survey}, making model evaluations more holistic and socially grounded.
By openly releasing our dataset, annotation schema, and reasoning templates, we further promote transparency, reproducibility, and community adoption.
From a deployment perspective, the inclusion of structured reasoning outputs facilitates integration into real-world clinical workflows. 
For instance, coupling an “eviction risk” flag with a concise model-generated rationale could help social workers and care teams quickly interpret and act upon risk signals. 
With appropriate clinical validation, such NLP-based surveillance tools could be integrated into existing national infrastructures, such as the U.S. Veterans Affairs data ecosystem, where they would complement structured resources like the Homelessness Management Information System (HOMES) and enhance population-level dashboards such as USVETS, as demonstrated by our previous nationwide study on NLP-identified eviction rates among over 7 million veterans~\cite{tsai2025eviction}.
Other studies have shown that enriching structured fields with NLP-derived factors improves the identification of at-risk patients~\cite{guevara2024large}.
In this context, SynthEHR-Eviction could help operationalize timely referrals to social services, housing aid, or financial counseling—interventions that may avert escalation to homelessness or emergency medical crises.
By enabling early and interpretable surfacing of eviction risk, SynthEHR-Eviction bridges a critical gap between clinical documentation and actionable care pathways. 
The structured rationales it produces not only improve transparency but also build clinician trust, two requirements for safe deployment in health systems.
Looking forward, we envision this work as a stepping stone toward broader SDoH-focused informatics innovations, including fine-grained ICD-Z code modeling, hybrid symbolic–neural architectures for social reasoning, and equitable NLP systems tailored to population health.
Ultimately, SynthEHR-Eviction demonstrates how modern LLM pipelines can be adapted to low-resource, high-impact clinical domains, advancing the broader agenda of embedding social medicine into the digital infrastructure of healthcare—a direction that is increasingly central to both health equity and precision public health.

\section{Methods}
\label{methods}

As illustrated in Figure~\ref{fig:eviction_annotation_design}, our approach focuses on developing a highly effective and reusable pipeline for multi-class annotation, which can be adapted to other SDoH annotation tasks.
In this study, eviction serves as the primary experimental focus, showcasing the pipeline's ability to generate high-quality annotations across multiple eviction-related categories.
The system is divided into three primary components: a training process for DSPy, data augmentation pipeline, and annotation system. 
The training process utilizes a small-scale human-validated dataset to optimize the DSPy program; the data augmentation pipeline generates labeled samples from raw clinical records; and the annotation system applies the trained models to classify augmented data and provide reasoning.




\subsection{Data Augmentation Pipeline}

\subsubsection{Raw Note Extraction}
We selected over 30,000 discharge notes from the MIMIC-III and MIMIC-IV databases. 
The selection criteria focused on notes that contained a "social history" section. This was based on our observation that most descriptions of patients' social circumstances—such as housing, financial status, and other social determinants—were concentrated within this part of the discharge notes. We then extracted the "social history" section from each note, using it as the raw note input for further augmentation.

\subsubsection{Data Construction}

\begin{figure}
    \centering
    \includegraphics[width=1\linewidth]{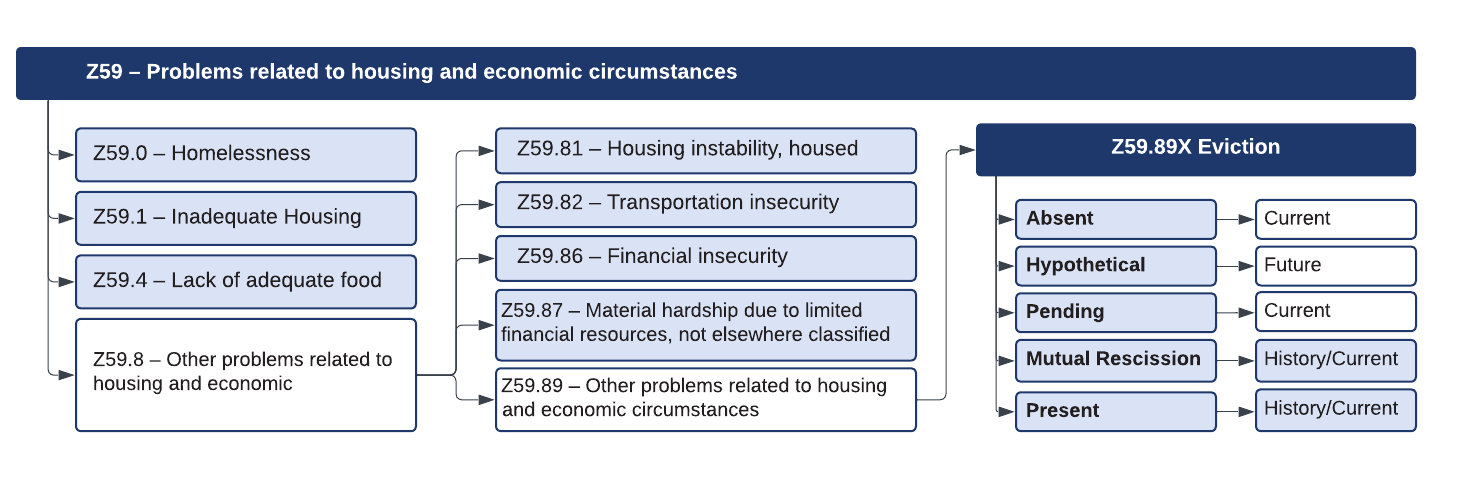}
    \caption{Relationships between Eviction and  other housing and economic challenges, providing a comprehensive framework for understanding its broader implications within the SDoH context.}
    \label{fig:eviction_classes}
\end{figure}

Based on SDoH data with ICD-10-CM Z Codes~\cite{cms_sdoz2023}, we categorize eviction as a new SDoH, classified under Z59.89, which addresses problems related to housing and economic circumstances. This classification positions eviction alongside other related classes under the Z59 category, as illustrated in Figure~\ref{fig:eviction_classes}. 

We classify eviction-related categories into 7 specific ones based on two key factors: presence (Absent, Hypothetical, Present, Pending, Mutual Rescission (MR)) and period (Current, History, Future)\cite{yao2023eviction}:
\begin{itemize}
\item Eviction Absent: No history or indication of eviction.
\item Eviction Present History, Eviction Present Current: Eviction has already been completed. "Current" refers to ongoing eviction impacts within the current natural year, not related to eviction. "History" refers to past without a specific time or that occurred more than one year ago.
\item Eviction Hypothetical: Refers to possible eviction based on hypothetical scenarios.
\item Eviction Pending: Eviction has been noticed but is not yet completed.
\item Eviction MR History, Eviction MR Current: MR is a legal agreement between the landlord and tenant to terminate a lease.
\end{itemize}

In addition to eviction-related categories, we included tier 1 categories: Homelessness (Z59.0), Inadequate housing (Z59.1) and Lack of adequate food (Z59.4). We also incorporated tier 2 categories including Housing instability (Z59.81), Transportation insecurity (Z59.82), Financial insecurity (Z59.86), and Material hardship (Z59.87).

In total, we defined 14 classes. For each class, experts provided precise definitions and a few-shot examples to guide data augmentation. The detailed definitions and examples for each category can be found in the Appendix Table~\ref{tab:definition_table_evi} and~\ref{tab:definition_table_nonevi}.

\begin{figure}
    \vspace{-15mm}
    \centering
    \includegraphics[width=1\linewidth]{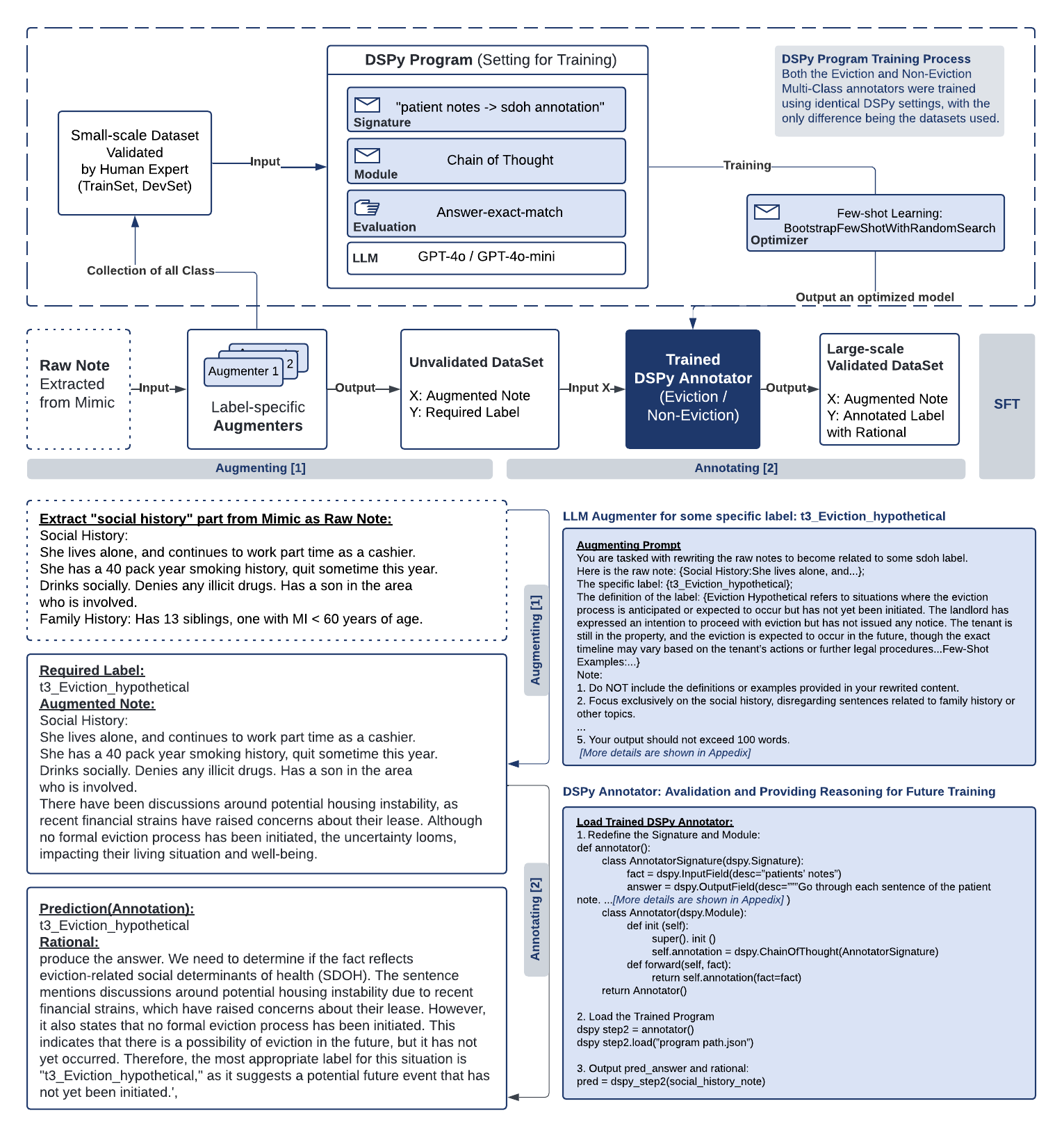}
    \vspace{-8mm}
    \caption{A complete DSPy-based system for training annotators to identify potential eviction cases from clinical notes. \textbf{Upper Section (Training Process)}: A small-scale human-validated dataset feeds into the DSPy Program training process, which employs GPT-4 or GPT-4o-mini models. The framework uses Chain of Thought reasoning and answer-exact-match evaluation, with few-shot learning optimization to produce a trained annotator model. \textbf{Lower Section (Pipeline Flow)}: \textbf{Augmentation Stage [1]}: Raw clinical notes extracted from MIMIC database undergo augmentation through label-specific augmenters to be the augmented notes dataset. \textbf{Annotation Stage [2]}: The trained DSPy Annotators (supporting Eviction/Non-Eviction classification separately) processes these augmented notes, outputting a large-scale validated dataset with annotated labels and accompanying rationales. \textbf{Bottom Section (Examples)} demonstrates a sample augmentation process for the t3\_Eviction\_hypothetical label and a detailed DSPy Annotator implementation and output format. \textbf{Quality Control}: The example shown represents a case where the annotated label matches the required label from augmentation, allowing direct inclusion in the final validated dataset. However, the system also handles annotation inconsistencies: If the annotated label differs from the required label, the system performs two additional annotation passes. When all three annotation outputs are consistent, the case is included in the final dataset with the annotated label (even if different from the original augmented label). If annotation outputs remain inconsistent, the case is discarded from the current validation cycle and the augmented note is returned to the raw note pool for potential future use. The entire framework maintains consistent settings across both Eviction and Non-Eviction Multi-Class annotators, with the only difference being the specific datasets used for training and test.}
    \label{fig:eviction_annotation_design}
\end{figure}

\subsubsection{Augmenter Construction Process}

As shown in Algorithm~\ref{alg:augmenter_pipeline}, we systematically developed an augmenter for each SDoH class. An augmenter in our context is essentially an LLM enhanced with an optimized prompt. Once the prompt is fully optimized, the augmenter takes raw notes as input and generates augmented notes that accurately reflect the target SDoH class.

For every SDoH class, the process starts by defining the SDoH category and providing raw notes to guide the LLM in generating relevant augmented notes.

\textbf{HITP Evaluation:} The output generated by the augmenter in each iteration undergoes human verification. Human experts verify if each output note accurately reflects the target SDoH. Verified notes are added to the augmented dataset, while incorrect outputs are logged with feedback.

\textbf{Assessment of Performance:} Accuracy is calculated as the proportion of correct annotations out of total outputs. If accuracy exceeds 90\%, the augmenter is deemed successful for that class. For augmenters not meeting the accuracy threshold, incorrect cases and feedback are returned to the LLM. The feedback is used to refine the prompt. The updated prompt is re-evaluated, and the cycle continues until the accuracy criterion is satisfied.

We conducted three optimization rounds to finalize augmenters for all 14 SDoH classes. For classes with clear distinctions and well-defined criteria (e.g., Tier 1 and Tier 2 SDoH categories), most augmenters reached the desired accuracy within a single iteration. For eviction-related classes, which involved finer granularity and ambiguous definitions, multiple iterations were required. This iterative process significantly improved the accuracy and reliability of the augmenters.

Each finalized augmenter demonstrated the capability to consistently generate augmented notes reflecting the target SDoH class. The optimized prompts for all classes are available in the Appendix.

\subsection{DSPy Annotator Setting and Training Process}

Inspired by the systematic survey of APO techniques~\cite{ramnath2025systematic}, the practical implementation of APO via DSPy~\cite{khattab2024dspy}, and our prior work demonstrating structured prompt optimization in healthcare settings~\cite{yao2024clinicians}, we design our annotator module as a critical component to ensure precise and efficient classification of notes into multiple SDoH categories.

\subsubsection{Annotator Setup and Design}
The dataset contains overlapping categories, particularly between Tier 1, Tier 2, and eviction-related classes. For instance, eviction can lead to homelessness or financial insecurity can lead to eviction, causing ambiguity in multi-class annotation. To address these overlaps, we first used a binary classification to determine whether content is related to eviction, and then divided the dataset into two groups, corresponding to three separate annotators, as outlined in Algorithm~\ref{alg:annotate_pipeline}.
The second annotator handles notes related to eviction-specific categories, including the following 7 labels: Eviction Absent, Eviction Present History, Eviction Present Current, Eviction Hypothetical, Eviction Pending, Eviction MR History, Eviction MR Current.
The third annotator manages notes that do not directly pertain to eviction, covering the remaining 7 labels: Homelessness, Inadequate Housing, Lack of Adequate Food, Housing Instability, Transportation Insecurity, Financial Insecurity, and Material Hardship.
This division ensures better class separation and significantly improves annotation accuracy.

\subsubsection{DSPy Training Workflows}
\begin{table}[h!]
\centering
\begin{tabular}{p{3cm}|p{13.5cm}}
\hline
\textbf{Component} & \textbf{Description} \\ \hline
\textbf{Signature} & 
Defines input-output behavior for DSPy modules.

Input: Patient Social History Note, Output: SDoH Annotation. \\ \hline
\textbf{Modules} & 
Each built-in module abstracts a prompting technique to handle signature.

Use \textbf{Chain-of-Thought (CoT) module} being step-by-step inference for accurate SDoH annotations. \\ \hline
\textbf{Evaluation Metric} & 
Uses \textbf{Answer-Exact-Match} to evaluate output correctness. Ensures that each predicted label aligns precisely with the ground truth.\\ \hline
\textbf{Optimizer} & 
A DSPy optimizer is an algorithm that can tune the parameters of a DSPy program (i.e., prompts and/or LM weights) to maximize the metrics.

We use \textbf{BootstrapFewShotWithRandomSearch}, which is building datasets for our modules and using them to finetune the LM weights in our system to iteratively select and evaluate example sets to optimize prompts. \\ \hline
\end{tabular}
\caption{Overview of DSPy Design in Annotation Pipeline}
\label{tab:dspy_pipeline}
\end{table}

The training process (upper section of Figure~\ref{fig:eviction_annotation_design}) is founded on a small-scale human-validated dataset, annotated by domain experts and divided into training and development sets. Our DSPy program employs either GPT-4 or GPT-4o-mini as the underlying language models, implementing some key components shown as Table~\ref{tab:dspy_pipeline}:

The output of the training process is an optimized DSPy model capable of accurately identifying potential eviction situations and providing corresponding reasoning. Any notes are input into this annotation tool to produce structured annotation results (annotation and detail reasoning), ensuring that the output dataset is high-quality, well-structured, and ready for downstream analysis.

\subsubsection{Trainset and Devset}
For each DSPy program, the trainset consists of data generated by the augmenter, specifically GPT-rewritten notes that have been verified by human evaluators. The trainset is carefully balanced, with 8 examples for each label category to ensure equal representation across all classification targets.

For the devset, we designed a more comprehensive evaluation to ensure the model's robustness and performance, with each label category represented by 48 examples. The devset includes three groups of data:

\begin{itemize}
\item Synth clinical notes: Similar to the trainset, this group consists of notes generated by GPT-4o-mini and verified by human experts, used to test the model's ability to generalize to data created by the augmenter.
\item MIMIC-IV~\cite{johnson2020mimic} clinical notes: To assess the model's performance on real-world data, we included raw notes from the MIMIC dataset. These notes primarily cover tier 1 and tier 2 SDoH categories but do not initially contain any references to eviction. To introduce eviction-related content, we first used a set of keywords (listed in Table~\ref{tab:keywords}) to filter the raw notes. For cases where this filtering process did not yield sufficient results, we employed human experts to rewrite a subset of these raw notes. The rewriting process was done carefully, ensuring that the original structure and characteristics of the notes were preserved while selectively adding eviction information.
\item PMC-Patient~\cite{zhao2023large} case reports: The third group comprises raw notes from PMC articles, following a similar strategy as the MIMIC raw notes. These notes were rewritten by humans to include targeted eviction-related scenarios without altering the inherent structure and characteristics of the original text.
\end{itemize}

By testing the model across these three distinct groups, we ensure that it can effectively handle both augmented and real-world data, as well as notes that have been selectively modified to introduce eviction scenarios. This diverse evaluation helps validate the model's performance and generalizability across various SDoH contexts.

\subsection{Leveraging Augmenters and Annotators for Fine-Tuning Open-Source Models}

Our fine-tuning process leverages the augmenter-annotator pipeline, which is instrumental in generating large volumes of high-quality labeled data. Label-specific augmenters were used to generate augmented notes tailored to the target labels. Each augmenter outputs synthetic data related to a specific SDoH category. The output from the augmenters was further annotated by the multi-label annotator. This annotator assigns the correct label along with corresponding reasoning.

In our fine-tuning process, we iteratively improved the model's performance through two phases. Initially, we exclusively used GPT-augmented sentences, which performed well on shorter texts but showed limited generalization on PMC long notes, highlighting the need for more diverse data. In the second phase, we enriched the training set by adding 30\% PMC long notes, significantly enhancing the model's ability to handle complex inputs. 

For the selection of open-source models, we conducted extensive experiments with multiple model architectures. We fine-tuned various models including BERT-based models (bert\_base\_cased-FT, biobert-v1.1-FT), domain-specific models (Bio\_ClinicalBERT-FT), and their SBDH-adapted variants (sdh\_bert\_base\_cased-FT, sdh\_biobert-v1.1-FT, etc.). We also explored LLMs such as LLaMA-3.1-8B-FT, LLaMA-3.2-3B-FT, Qwen2.5-7B-FT, and Qwen2.5-3B-FT, along with their SBDH-adapted versions. While all models demonstrated improvements after fine-tuning, the larger models (particularly the 7B and 8B variants) showed superior capacity for generalization and performance compared to the 3B-level models, which had more limited gains. This comprehensive approach allowed us to identify the most effective model architectures for our specific task.

\begin{table}[ht]
\centering
\footnotesize
\begin{tabular}{l|cc|c|cccc}
\hline
\textbf{Label} & \textbf{DSPy-train (Synth)} & \textbf{DSPy-eval (Synth)} & \textbf{SFT-train+eval} & \textbf{Test-total} & \textbf{Test-Synth} & \textbf{Test-Mimic} & \textbf{Test-PMC} \\
\hline
Eviction\_absent & 8 & 12 & 500 & 20 & 20 &  &  \\
Eviction\_hypothetical & 8 & 12 & 750 & 48 & 20 & 20 & 8 \\
Eviction\_mr\_current & 8 & 12 & 750 & 48 & 20 & 20 & 8 \\
Eviction\_mr\_history & 8 & 12 & 750 & 48 & 20 & 20 & 8 \\
Eviction\_pending & 8 & 12 & 750 & 48 & 20 & 20 & 8 \\
Eviction\_present\_current & 8 & 12 & 750 & 48 & 20 & 20 & 8 \\
Eviction\_present\_history & 8 & 12 & 750 & 48 & 20 & 20 & 8 \\
\hline
Eviction\_total & 56 & 84 & 5000 & 308 & 140 & 120 & 48 \\
\hline
Homelessness & 8 & 12 & 450 & 20 & 20 & &  \\
InadequateHousing & 8 & 12 & 450 & 48 & 20 & 20 & 8 \\
LackOfAdequateFood & 8 & 12 & 450 & 48 & 20 & 20 & 8 \\
FinancialInsecurity & 8 & 12 & 450 & 48 & 20 & 20 & 8 \\
HousingInstability & 8 & 12 & 450 & 48 & 20 & 20 & 8 \\
MaterialHardship & 8 & 12 & 450 & 48 & 20 & 20 & 8 \\
TransportationInsecurity & 8 & 12 & 300 & 48 & 20 & 20 & 8 \\
\hline
Non-Eviction\_total & 56 & 84 & 3000 & 308 & 140 & 120 & 48 \\
\hline
\end{tabular}
\caption{Data statistics across training and evaluation splits for APO (for close-source models) and SFT (for open-source models).}
\label{tab:data_stats}
\end{table}

\subsection{Experimental Settings}

\subsubsection{Main Experimental Setup}
We conducted a comprehensive evaluation across three sequential steps (Step 1: Binary Classification, Step 2: Eviction Multi-Class Classification, and Step 3: Non-Eviction Multi-Class Classification). For each model, we performed 5 independent runs to calculate 95\% confidence intervals, ensuring the statistical reliability of our results.

For language models like GPT-4o and GPT-4o-mini, we initially used temperature$=0$ for the annotation task to minimize uncertainty. However, to compute reliable confidence intervals, we performed 4 additional runs with temperature$=0.5$. 
For models utilizing dspy frameworks and other LM/LLM architectures, we maintained consistency by using different random seeds across the 5 runs. Throughout all experiments, we kept the fine-tuning process consistent, with identical fine-tuning parameters across all models and training set sizes.
To ensure compliance with MIMIC-IV data usage requirements~\footnote{\url{https://physionet.org/news/post/gpt-responsible-use}}, we exclusively used the Azure OpenAI service for GPT-4o and GPT-4o-mini, with human review explicitly disabled, following PhysioNet’s guidelines for responsible use of credentialed data.
We introduce the detailed experimental settings in the appendix.

Each model was evaluated on three distinct development sets (Synth, Mimic, and PMC), with the average performance calculated by simply taking the arithmetic mean across these three datasets. This approach allows us to assess model generalization across different data distributions.

The overall performance metric was derived from the best-performing model configuration for each step. Importantly, we consider a prediction correct only when Step 1 (Binary Classification) is correct, and either Step 2 (Eviction Multi-Class Classification) or Step 3 (Non-Eviction Multi-Class Classification) is also correct for the same clinical note. This evaluation approach reflects the cascading nature of our classification task, where Step 1 determines whether a note contains eviction class, while Steps 2 and 3 classify non-overlapping subsets of notes.

This evaluation framework ensures that models must perform well across the entire classification pipeline to achieve high overall scores, providing a realistic assessment of their practical utility in clinical document analysis.

\subsubsection{Training Data Size Impact Experiment} We tested a diverse set of models including traditional LMs (BERT variants) and LLMs (Llama and Qwen families) on both Eviction and Non-Eviction Multi-Class Classification. We used five different training set sizes (200, 1000, 3000, 5000, and 10000 samples) to establish clear performance trends. As the dataset size increased, the number of examples in each class increased proportionally. The datasets for BERT-based models contained only labels without reasoning, while LLM datasets included both labels and reasoning. Each training set was run only once, with consistent fine-tuning parameters across all experiments. 

\subsubsection{Reasoning in Trainset Experiment} 
We trained LLMs (Llama-3.1-8B, Llama-3.2-3B, Qwen2.5-7B, and Qwen2.5-3B) using two distinct training approaches: 
(1) With Reasoning: Models were trained on datasets containing both classification labels and explicit reasoning steps.
(2) Without Reasoning: Models were trained on identical datasets but with only classification labels (reasoning steps removed).
All models were evaluated on both Eviction and Non-Eviction Multi-Class tasks using Micro-F1 scores as the primary performance metric.
All fine-tuning hyperparameters were kept consistent between the with-reasoning and without-reasoning training conditions to isolate the specific impact of reasoning annotations.


    




\subsection{Metrics}
Following Yao et al. (2023)~\cite{yao2023eviction}, we calculate the confusion matrix C for both tasks, which are True Positives (TP), True Negatives (TN), False Positives (FP), and False Negatives (FN) for each class.
We use both \textbf{Micro-F1} and \textbf{Macro-F1} in our evaluation. 
We also report the Matthews Correlation Coefficient (\textbf{MCC}) score, a metric that yields a high score only if the prediction yields good results in all four confusion matrix categories (TP, TN, FP, and FN). 


\section{Data Availability}

Synth : \url{https://huggingface.co/datasets/youxiazhao/eviction-devset-synth} \\
PMC-Patients: 
\url{https://huggingface.co/datasets/youxiazhao/eviction-devset-pmc} \\
Only individuals who have signed the MIMIC Data Use Agreement may contact us to obtain the MIMIC-based version of the dataset.

\section{Code Availability}

The code is publicly available on Github: \url{https://github.com/youxiazhao/sdoh_annotation}

\section{Acknowledgements}
Research reported in this study was supported by the National Center on Homelessness Among Veterans (NCHAV) and by the National Institutes of Health (NIH) under award number 1R01NR020868, and 1I01HX003711-01A1.
This study was also in part supported by NIH under award numbers R01DA056470-A1 and 1R01AG080670-01, and by the U.S. Department of Veterans Affairs (VA) Health Systems Research.
The content is solely the responsibility of the authors and does not necessarily represent NIH, VA, or the US government. 


\section{Author contributions statement}

Z.Y., H.Y. and J.T. designed the study. 
Z.Y. and Y.Z. performed the data collection. 
Z.Y. and Y.Z. implemented the code, and conducted
experiments.
Z.Y., Y.Z., J.T., A.M. and H.Y. drafted the manuscript. 
H.Y. supervised the study. 
D.L. and E.D. performed manual annotation and human evaluation.
All authors contributed to the research discussion,
manuscript revision, and approval of the manuscript for submission.

\section{Competing Interests}

The authors declare no competing interests.

\section*{Declaration of generative AI and AI-assisted technologies in the writing process}

During the preparation of this work, the author(s) used ChatGPT to improve the language and readability. Following the use of this tool, the author(s) carefully reviewed and edited the content as necessary and take full responsibility for the final version of the manuscript.

\bibliography{sample}

\newpage
\section*{Appendix}
\label{appendix:appendix}

\begin{table}[ht]
\footnotesize 
\centering
\begin{minipage}{0.42\textwidth}
\centering
\begin{tabular}{l|c|c|c}
\toprule
\multicolumn{4}{c}{\textbf{Step 2: Eviction Multi-Class}} \\
\midrule
Training Set & GPT-4o-APO & Llama-3.1-8B-FT & Qwen2.5-7B-FT \\
\midrule
100\% Synth & 0.835(0.692) & 0.797(0.606) & 0.800(0.558)\\
70\% Synth 30\% PMC & 0.915(0.929) & 0.913(0.888) & 0.912(0.860) \\
\bottomrule
\end{tabular}
\end{minipage}
\hfill
\begin{minipage}{0.42\textwidth}
\centering
\begin{tabular}{c|c|c}
\toprule
\multicolumn{3}{c}{\textbf{Step 3: Non-Eviction Multi-Class}} \\
\midrule
GPT-4o-APO & Llama-3.1-8B-FT & Qwen2.5-7B-FT \\
\midrule
0.863(0.789) & 0.870(0.800) & 0.864(0.793) \\
0.911(0.922) & 0.929(0.942) & 0.925(0.944) \\
\bottomrule
\end{tabular}
\end{minipage}
\caption{Comparison of model performance across different training data compositions.}
\label{tab:training_composition_table}
\end{table}

\paragraph{Training Set Composition} Our initial training approach utilized purely synthetic data for model training. However, this configuration demonstrated suboptimal performance on PMC note evaluations, revealing limitations in the synthetic data's ability to capture the nuances of real-world medical documentation. To address this limitation, we modified our training set composition by replacing 30\% of synthetic data with actual PMC notes. This hybrid approach yielded substantial improvements across different experimental settings. As shown in Table \ref{tab:training_composition_table}, the introduction of PMC data led to an increase of 0.237 in Micro-F1 score on the PMC development set in our DSPy-based experiments, while achieving an overall average improvement of 0.08 across all evaluation metrics. Similarly, in the LLM fine-tuning experiments, such as with Qwen2.5-7B, the hybrid training set demonstrated even more pronounced benefits, with performance on PMC-specific tasks improving by 0.302 while overall performance increased by 0.112.

\paragraph{Fine-tuneing Settings}
Our fine-tuning methodology was tailored to each model architecture to optimize performance while maintaining computational efficiency. For large language models (LLaMA and Qwen variants), we employed Unsloth~\footnote{\url{https://github.com/unslothai/unsloth}}, a parameter-efficient fine-tuning framework that enables significantly faster training and inference compared to conventional methods.

For LLaMA and Qwen models, we implemented the following configuration using Unsloth:
\begin{itemize}
    \item Maximum sequence length of 1024 tokens with 4-bit quantization (load\_in\_4bit=True) to reduce memory usage
    \item LoRA (Low-Rank Adaptation) with rank (r) of 16 and alpha value of 16 Target modules included query, key, value projections ('q\_proj', 'k\_proj', 'v\_proj', 'o\_proj', 'gate\_proj', 'up\_proj', 'down\_proj')
    \item LoRA dropout set to 0 and bias configuration set to 'none' for optimized performance
    \item Learning rate of 2e-4 with a linear scheduler
    \item Training for 2 epochs with gradient accumulation steps of 4
    \item Warmup steps of 5 and weight decay of 0.01
    \item Per-device batch size of 2 with AdamW 8-bit optimizer
\end{itemize}

Our training dataset was split into training and test sets with an 80/20 ratio (test\_size=0.2), using a seed of 42 for reproducibility. For data formatting, we employed a chat template structure with a system prompt instructing the model to act as a healthcare annotator specifically focused on eviction-related content. This approach helped align the model's responses with our annotation requirements.

For BERT-based models (bert\_base\_cased, biobert, etc.), we utilized the standard fine-tuning approach from Hugging Face's Transformers library. These models were trained with the same learning rate (2e-4) and epochs (2) as the LLMs for consistency, but with appropriate modifications to accommodate their architectural differences. The fine-tuning was performed on a Tesla T4 GPU with 14.748 GB of memory.

The hyperparameters for each model were determined through preliminary experiments, with optimal configurations selected based on Micro-F1 scores for our target SBDH categories. This comprehensive approach allowed us to identify the most effective model architectures while maintaining computational efficiency through parameter-efficient fine-tuning techniques.

\begin{figure}[!ht]
    \centering
    \includegraphics[width=1\linewidth]{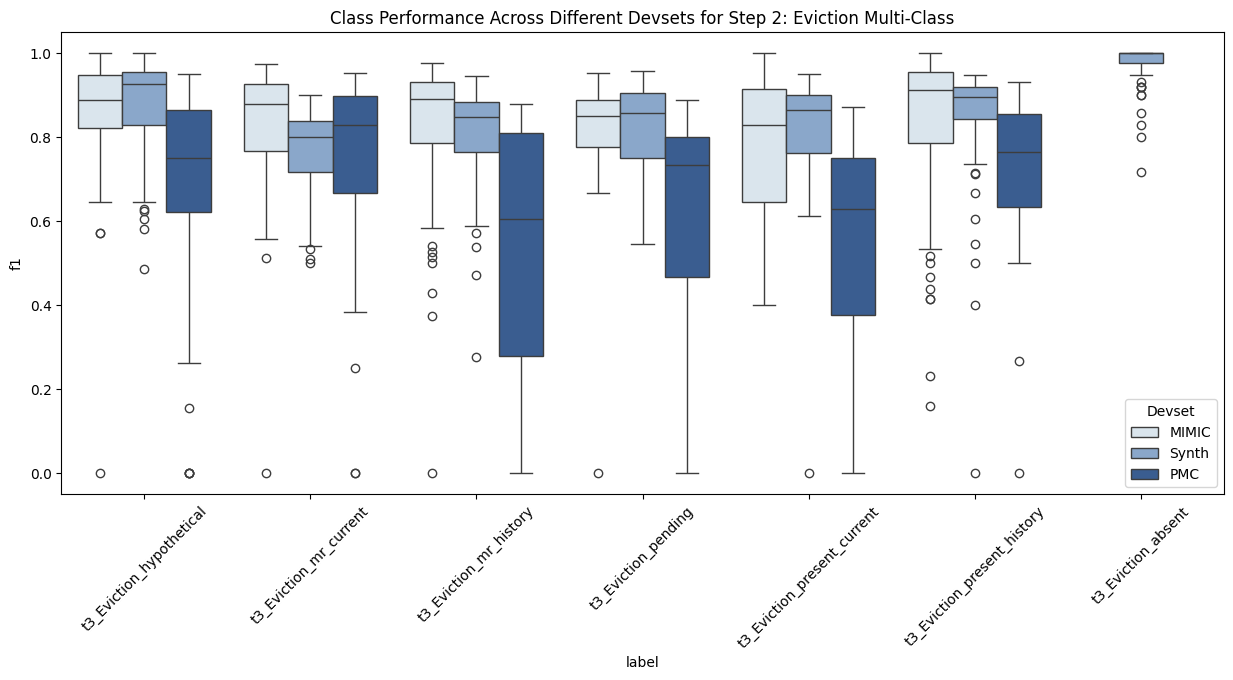}
    \caption{Class Performance Across Different Devsets for Step 2: Eviction Multi-Class. Box plot revealing $F1$ scores across three datasets (Synth, MIMIC, and PMC) for all evaluated models, including GPT-4o-APO, GPT-4o-mini-APO, fine-tuned Qwen, Llama, SBDH-LLM, BERT, Clinical-Bert and SDoH-BERT. Each model was evaluated across 5 independent runs, revealing both inter-dataset variations and substantial performance differences between eviction-related labels}
    \label{fig:class_boxplot_step2}
\end{figure}

\paragraph{Performance Variations Across Different Devsets}
Our comprehensive analysis compared performance across three distinct datasets: Synth, MIMIC, and PMC in the Eviction Multi-Class task (Figure~\ref{fig:class_boxplot_step2}). The Synth dataset consistently demonstrated the most stable performance characteristics, exhibiting high and uniform $F1$ scores across eviction classes. Similarly, the MIMIC dataset showed minimal variability, with stable and reliable performance across most classes. In contrast, the PMC dataset demonstrated the most significant performance heterogeneity, characterized by the widest performance ranges and lower median scores. This performance degradation aligns with the trends observed in Table~\ref{tab:main_table}, where models exhibited substantially lower micro-F1 scores on PMC data compared to Synth and MIMIC datasets. These results highlight notable inter-dataset variations, particularly emphasizing the challenges posed by PMC notes.

\paragraph{Interpretation of Inter-Dataset Variability}
The performance discrepancies between datasets, particularly the pronounced heterogeneity in PMC results, can be attributed to both structural differences in the clinical notes and model limitations. Specifically, PMC notes tend to be significantly longer and more detailed, leading to truncation due to our 512-token input limit, which likely eliminates crucial contextual information. Moreover, unlike the Synth and MIMIC datasets, where eviction-related information is often explicitly stated and follows a structured format, PMC notes embed relevant details within complex, narrative clinical discussions. This structural complexity complicates the model's ability to generalize, especially when it has been primarily trained on more structured datasets.

\begin{table}[htbp]
    \centering
    \footnotesize
    \caption{Eviction-related Class Definitions}
    \label{tab:definition_table_evi}
    \begin{tabular}{p{2.7cm}|p{14cm}}
        \toprule
        \textbf{SDoH Class} & \textbf{Definition} \\
        \midrule

        Eviction\_present\_current & 
        1. Eviction Present refers to the state where the eviction process has already been fully concluded, and the tenant has been legally removed from the property. All legal proceedings, such as notices and hearings, have been completed, and the tenant no longer has access to the property.
        \newline
        - Consideration: When generating the note, it should mention that the eviction has already taken place, and the tenant is no longer living at or has access to the property.  
        \newline
        2. Current refers to events related to eviction that happened currently or recently, with specificity about the timing. This includes cases where the eviction event occurred in the current year (e.g., "this year," "last month", "last week", "several months ago," or "a few months ago", "recently"...).  
        \newline
        - Consideration: The note should mention that the event was in the current year, providing some specific time reference. \\
        \hline

        Eviction\_present\_history & 
        1. Eviction Present refers to the state where the eviction process has already been fully concluded, and the tenant has been legally removed from the property. All legal proceedings, such as notices and hearings, have been completed, and the tenant no longer has access to the property.
        \newline
        - Consideration: When generating the note, it should mention that the eviction has already taken place, and the tenant is no longer living at or has access to the property.
        \newline
        2. History refers to events related to eviction that happened in the past, but with less specificity about the timing. This includes cases where the eviction event occurred in the distant past (e.g., "last year," "several years ago," or "a few years ago"), or where the mutual rescission agreement or eviction action itself isn't explicitly stated.
        \newline
        - Consideration: The note should mention that the event was in the past, providing some general time reference, but without specifying the exact date or events.\\
        \hline

        Eviction\_pending & 
        Eviction Pending refers to a situation where eviction proceedings have been initiated but are not yet complete. The tenant is still in the property, and there is still an opportunity for negotiation, remediation, or resolution before a final court decision or physical removal occurs. This state indicates that while the eviction process has started, the outcome is still undecided, and there is a potential for the tenant to address the issue and avoid eviction.
        \newline
        - Distinction from Completed Eviction: Unlike a completed eviction where the tenant has already been removed, eviction-pending indicates that the tenant has received a notice, but there has been no final court order or physical removal. The tenant may negotiate, pay overdue rent, or comply with other conditions to potentially stop the eviction process. 
        \newline
        Few-Shot Examples:
        \newline
        - “The tenant received an eviction notice recently, but negotiations with the landlord to pay overdue rent are still ongoing.”  
        \newline
        - “A few months ago, the landlord filed for eviction due to nonpayment, but the case is still pending in court, giving the tenant a chance to settle.”
        \newline
        - “Currently, the tenant is under an eviction notice but is working with a housing advocate to resolve the issue before the court date.” \\
        \hline

        Eviction\_hypothetical & 
        Eviction Hypothetical refers to situations where the eviction process is anticipated or expected to occur but has not yet been initiated. The landlord has expressed an intention to proceed with eviction but has not issued any notice. The tenant is still in the property, and the eviction is expected to occur in the future, though the exact timeline may vary based on the tenant's actions or further legal procedures.
        \newline
        - Time Frame: "hypothetical" in this context refers to actions or plans related to eviction that are expected to occur soon. This could include statements like "in the coming weeks," "next month," or "soon," indicating that the eviction is planned but not yet happend.
        \newline
        Few-Shot Examples:
        \newline
        - “The landlord has given the tenant a final warning, and eviction proceedings are expected to start next month if the rent is not paid.”
        \newline
        - “The tenant has been notified that they must vacate the premises in the coming weeks due to repeated violations of the lease.”
        \newline
        - “Eviction is planned for the near future, as the landlord has expressed intent to reclaim the property due to nonpayment.”
        \newline
        - “The landlord mentioned that they will file for eviction soon if the tenant does not comply with the notice to remedy the situation.”
        \newline
        - “Next week, the landlord plans to serve an eviction notice due to ongoing illegal activities on the property.” \\
        \hline

        Eviction\_mr\_current & 
        1. Mutual Rescission (mr) refers to a specific legal agreement in which both the landlord and tenant agree to terminate the lease early. This occurs after eviction proceedings have started, but the eviction process has not yet reached its final stage. As a result of this agreement, the tenant voluntarily vacates the rental property, and the eviction process is stopped before completion. All necessary legal proceedings are concluded, and the tenant no longer has access to the property. 
        \newline
        - Consideration: When generating the note, it should mention that the Mutual Rescission has already taken place, the eviction process had already stopped, and the tenant is no longer living at or has access to the property.
        \newline
        2. Current means that the agreement or action related to mutual rescission is mentioned within a recent period (e.g., "a few months ago," "recently," "this year"). \\
        \hline
        
        Eviction\_mr\_history & 
        1. Mutual Rescission (mr) refers to a specific legal agreement in which both the landlord and tenant agree to terminate the lease early. This occurs after eviction proceedings have started, but the eviction process has not yet reached its final stage. As a result of this agreement, the tenant voluntarily vacates the rental property, and the eviction process is stopped before completion. All necessary legal proceedings are concluded, and the tenant no longer has access to the property.
        \newline
        - Consideration: When generating the note, it should mention that the Mutual Rescission has already taken place, the eviction process had already stopped, and the tenant is no longer living at or has access to the property.
        \newline
        2. History refers to events related to eviction that happened in the past, but with less specificity about the timing. This includes cases where the Mutual Rescission occurred in the distant past (e.g., "last year," "several years ago," or "a few years ago"), or where the mutual rescission agreement or eviction action itself isn't explicitly stated.
        \newline
        - Consideration: The note should mention that the event was in the past, providing some general time reference, but without specifying the exact date or events. \\
        \hline

        Eviction\_absent & 
        The text clearly states "never evicted" or "no history of eviction". \\

        \bottomrule
    \end{tabular}
\end{table}

\begin{table}[!htbp]
    \centering
    \footnotesize
    \caption{Non-Eviction-related Class Definition}
    \label{tab:definition_table_nonevi}
    \begin{tabular}{p{2.7cm}|p{14cm}}
        \toprule
        \textbf{SDoH} & \textbf{Definition (/Examples)} \\
        \midrule
    Homelessness & An individual or family who lacks a fixed, regular, and adequate nighttime residence, such as those living in emergency shelters, transitional housing, or places not meant for habitation.

        Here are some few-shot examples:
        "...is homeless and lives in a shelter…",
        “...contact one of the homeless shelters....",
        “...found it difficult to secure housing and ended up living in his car…”,
        “...relying on friends and temporary shelters for support…”,
        “...is actively seeking employment and more permanent housing but has faced numerous obstacles…”,
        “living on the streets…”,
        “...living in a homeless encampment...”,
        “...couch surfing...” \\  
        \hline
        
        InadequateHousing & Inadequate housing is defined as an occupied housing unit that has moderate or severe physical problems (e.g., deficiencies in plumbing, heating, electricity, hallways, and upkeep). Examples of moderate physical problems in a unit include two or more breakdowns of the toilets that lasted more than 6 months, unvented primary heating equipment, or lack of a complete kitchen facility in the unit. Severe physical problems include lack of running hot or cold water, lack of a working toilet, and exposed wiring.

        Here are some few-shot examples:
        “...lives in an old apartment building with severe structural issues...”,
        “...live in an apartment that lacks a functioning heating system…”,
        “...a family, consisting of six members spanning three generations, lives in a cramped two-bedroom apartment…”,
        “...unsafe housing situation…”,
        “...unsanitary living conditions…”,
        “...polluted living environment…”,
        “...lead and toxic exposures in home…” \\ 
        \hline
        
        LackOfAdequateFood & Food insecurity is the limited or inadequate access to food because of insufficient money and other resources for food.
        Food security, at the individual, household, national, regional, and global levels [is achieved] when all people, at all times, have physical and economic access to sufficient, safe, and nutritious food to meet their dietary needs and food preferences for an active and healthy life. This definition suggests that food insecurity is the absence of one or more of these conditions.

        Here are some few-shot examples:
        “...lack the variety and nutrients... “,
        “...frequently goes hungry or eats whatever is available...”,
        “...no supermarkets...”,
        “...difficult to access better food sources outside the neighborhood...”,
        “...uses food pantries/soup kitchens for food…”,
        “...does not have stable food sources…”,
        “...lives in a food desert…” \\  
        \hline
        
        FinancialInsecurity & Economic insecurity can be defined as “the anxiety produced by the possible exposure to adverse economic events and by the anticipation of the difficulty to recover from them”. Examples could include a fear of unemployment, an expectation of a worsening financial situation, money mismanagement, or being financially exploited or a victim of financial scam.

        Here are some few-shot examples:
        “...has been having a lot of stress recently due to financial concerns.”,
        “...was very concerned about the financial burden of hospitalization, medications and potential surgery.”,
        “Rising living costs, including healthcare and housing, have made it difficult for them to cover their monthly expenses.”,
        “The irregular income and lack of benefits make it difficult to budget and plan for the future.”,
        “...has experienced several months of financial difficulty due to job loss.”,
        “...not managing money well…”,
        “...lacks financial literacy…”,
        “...mismanaging funds.”,
        “...lack of stable income…” \\  
        \hline
        
        HousingInstability & Housing instability is variably defined as having difficulty paying rent, spending more than 50\% of household income on housing, having frequent moves, living in overcrowded conditions, or doubling up with friends and relatives. 
        Unstably housed, housing insecure, or in a temporary housing situation. 
        At risk of being homeless or at imminent risk of being homeless.
        
        Here are some few-shot examples:
        “...has moved three times...”,
        “…fell behind on his rent payments...”,
        “...live together with three families to share...”,
        “...couch surfing…”,
        “...temporarily staying with friends/family…”,
        “...staying in a motel/hotel…”,
        “...in a temporary housing situation…”,
        “...at risk of losing their housing…” \\
        \hline
        
        MaterialHardship & Material hardships, defined as difficulty meeting basic needs such as food, housing or medical care, are common among low-income households.

        Here are some few-shot examples:
        “Their electricity was cut off because they couldn't make the payments.”,
        “...cannot afford to buy winter coats or shoes that fit correctly.”,
        “...cannot afford the necessary school supplies and textbooks for their three children.”,
        “...cannot afford to participate in health and wellness activities such as exercise classes, which are essential for managing his chronic conditions.” \\  
        \hline
        
        TransportationInsecurity & Transportation insecurity occurs when a person is regularly unable to get from place to place in a safe or timely manner because of a lack of resources. This can limit a person's access to work, school, medical care, social activities, and more.

        Here are some few-shot examples:
        “…lives in a rural area where there are no public transportation options. The nearest town with essential services such as grocery stores, schools, and healthcare facilities is 20 miles away.”,
        “Buses and trains are often delayed or overcrowded, making it difficult for her to get to work on time.”,
        “...can't afford the transportation fare, he and his children have to walk long distances, often in unsafe conditions.”,
        “...uses a wheelchair and often finds that public transportation in her city is not fully accessible.”,
        “...does not own a car…”,
        “...does not have bus passes…” \\
        
        \bottomrule
    \end{tabular}
\end{table}

\begin{table}[htbp]
    \centering
    \scriptsize
    \caption{Human Instructions / LLM Prompt}
    \label{tab:prompt}
    \begin{tabular}{p{2.3cm} p{13cm}}
        \toprule
        \textbf{Instruction / Prompt} & \textbf{Detail}\\  
        \midrule
        Augmenting Prompt & You are tasked with rewriting the raw notes to become related to some SDoH label.\newline
        Here is the raw note: ${raw\_notes}$\newline
        And the specific label: ${label}$\newline
        The definition of the label: ${definition}$\newline
        Note:\newline
        1. Do NOT include the definitions or examples provided in your rewrited content.\newline
        2. Focus exclusively on the social history, disregarding sentences related to family history or other topics.\newline
        3. The augmented notes should clearly reflect the label context, be contextually coherent, with varied and diverse expressions.\newline
        4. The augmented note should be a detailed description of a specific patient case that illustrates the application of the SDoH label. Focus on the unique circumstances, events, and actions related to this individual case. Avoid using general or broad descriptions of processes or procedures; instead, provide concrete details and examples that are directly relevant to the patient's situation.\newline
        5. Your output should not exceed 100 words.\newline
        Augmented Notes:\newline
        
        input\_variables=["raw\_notes", "label", "definition"]\\
        \midrule
        Annotation Prompt in Step 1: Binary Classification & Go through each sentence of the patient note. If a sentence reflects eviction-related social determinants of health (SDoH), assign the label "Yes", else annotate as label "No"\\
        \midrule
        Annotation Prompt in Step 2: Eviction Multi-Class & Go through each sentence of the patient note. If a sentence reflects eviction-related social determinants of health (SDoH), assign the most appropriate label from the following list: "t3\_Eviction\_absent", "t3\_Eviction\_present\_history", "t3\_Eviction\_present\_current", "t3\_Eviction\_pending", "t3\_Eviction\_mr\_history", "t3\_Eviction\_mr\_current", "Other". For status part, if no eviction in the history and in the future: "absent"; if eviction is completed: "present"; if eviction noticed but not completed: "pending"; if eviction might be happend in the future: "hypothetical"; if mutual rescission: "mr". For timeframe part when "present" or "mr" status, if it is happened within this natural year: "current". If not shown specific time or noticed a time before this natural year: "history". \\
        \midrule
        Annotation Prompt in Step 3:  Non-Eviction Multi-Class & Choose the most approperate label from "t1\_Homelessness", "t1\_InadequateHousing", "t1\_LackOfAdequateFood", "t2\_FinancialInsecurity", "t2\_HousingInstability", "t2\_MaterialHardship", "t2\_TransportationInsecurity", "Other".\newline
        't1\_Homelessness': An individual or family who lacks a fixed, regular, and adequate nighttime residence, such as those living in emergency shelters, transitional housing, or places not meant for habitation.\newline
        't1\_InadequateHousing': an occupied housing unit that has moderate or severe physical problems (e.g., deficiencies in plumbing, heating, electricity, hallways, and upkeep)\newline
        't1\_LackOfAdequateFood': is the limited or inadequate access to food because of insufficient money and other resources for food.\newline
        't2\_FinancialInsecurity': the anxiety produced by the possible exposure to adverse economic events and by the anticipation of the difficulty to recover from them. Examples could include a fear of unemployment, an expectation of a worsening financial situation, money mismanagement, or being financially exploited or a victim of financial scam.\newline
        't2\_HousingInstability': having difficulty paying rent, spending more than 50\% of household income on housing, having frequent moves, living in overcrowded conditions, or doubling up with friends and relatives.
        't2\_MaterialHardship': difficulty meeting basic needs such as food, housing or medical care, are common among low-income households.\newline
        't2\_TransportationInsecurity': occurs when a person is regularly unable to get from place to place in a safe or timely manner because of a lack of resources. This can limit a person's access to work, school, medical care, social activities, and more.\\
        \midrule
        Human Annotation for Augmentation Period& We need to check the quality of the GPT-generated notes to refine our prompts, and this requires manual verification—your role in this project. You'll review the entries, marking them 'True' if correct, and giving feedback if false.\newline
        We have 14 labels, (including "t1\_Homelessness", "t1\_InadequateHousing", "t1\_LackOfAdequateFood", "t2\_FinancialInsecurity", "t2\_HousingInstability", "t2\_MaterialHardship", "t2\_TransportationInsecurity", "t3\_Eviction\_absent", "t3\_Eviction\_present\_history", "t3\_Eviction\_present\_current", "t3\_Eviction\_pending", "t3\_Eviction\_mr\_history", "t3\_Eviction\_mr\_current"), which make up our initial data set, with 20 entries for each label. We've engaged three experts to ensure accuracy by cross-verifying the data. Each label is annotated by two different experts, so each expert will annotate over 200 entries.\newline
        If you are not sure the definitions of some label, you can check the document (Same as Table ~\ref{tab:definition_table_evi} and~\ref{tab:definition_table_nonevi}). \\
        \midrule
        Manual Rewrite Notes for Augmentation & We need your assistance to rewrite clinical notes from MIMIC to reflect specific social determinants of health (SDoH) classes related to eviction. For each note and assigned class/label, please modify the content to clearly demonstrate the corresponding eviction situation while preserving the original MIMIC documentation style. Your rewrites should maintain the clinical tone, formatting, and structure of the original notes (including headers, abbreviations, and documentation patterns typical in MIMIC), but incorporate eviction-related circumstances that align with the specified class (e.g., t3\_Eviction\_present\_history, t3\_Eviction\_pending, etc.). The goal is to create authentic-looking clinical documentation that can be used to augment our training data for eviction classification models.\newline
        If you are not sure the definitions of some label, you can check the document (Same as Table ~\ref{tab:definition_table_evi} and~\ref{tab:definition_table_nonevi}).  \\
        \midrule
        Manual Double-Check Rewritten Notes & We have some data entries related to eviction that were rewritten by medical students. However, since they are not experts, we need your help to double-check their correctness. This is the same as the task you’ve worked on before—just review and mark each entry as either True or False.\\ 
        \bottomrule
    \end{tabular}
\end{table}

\begin{table}[ht]
   \centering
   \footnotesize 
   \renewcommand{\arraystretch}{1} 
   \setlength{\tabcolsep}{2pt} 
   \begin{tabularx}{\textwidth}{l|X}
   \toprule
   \textbf{Training Set} & \textbf{Sample Note} \\
   \midrule
   Synth & The patient, a widowed immigrant living with her daughter, faces potential housing instability. Having lived in the country for about 50 years, her current housing situation may be threatened, necessitating discussions about mutual rescission to ensure she can voluntarily vacate her rental early if needed, which reflects her vulnerabilities related to social determinants of health. \\
   \midrule
   MIMIC & Social History:
+ Tob, 1.5 ppy X many years, no EtOH, was forced to remove from the rented house, has a 25yo
son \\
   \midrule
   PMC & Our 13-year-old patient is a Caucasian girl with an unremarkable medical history, yet her family history is significant for allergies and asthma. She initially presented, to another institute, with multiple pruritic facial skin lesions and a pruritic left intranasal lump. Apart from having two erythematous pruritic plaques in the left suborbital region and a yellowish pruritic lump occupying the left nasal vestibule, her physical examination proved to be insignificant without any lymphadenopathies or salivary gland enlargements. Consequently, a laboratory workup was conducted in addition to an excisional biopsy of one dermatologic lesion and a needle biopsy of the nasal lesion. Both biopsies exhibited nonspecific inflammation with granulation, necrosis, and no signs of malignancy. No specific diagnosis was made. The patient was started on hydrogen peroxide treatment for the skin lesions, which resolved completely with no recurrence. Simultaneously, another lump started growing in the right nasal vestibule. Suspecting an inflammatory etiology, she was started on oral prednisolone 1 mg/kg/day by mouth twice a day. Despite therapy, these nasal lumps continued growing. Hence, prednisolone was discontinued after 10 days of therapy to start an indomethacin trial of 1.5 mg/kg/day by mouth twice a day for 14 days, which was also discontinued due to its inefficacy. Then a decision to perform surgery was taken, and the presurgical computed tomography (CT) scan revealed bilateral soft tissue masses arising from the right and left nasal vestibules. Although she had undergone many surgical attempts to remove the lumps, none of them succeeded and both lumps flared in size. After a couple of months, our patient presented to our institution with bilateral nasovestibular lumps; they were massive in size, occluding nasal entrance and protruding outside the nose (Fig. ). We did an extensive laboratory workup to exclude any comorbidities (Table ). We did a fine-needle aspiration (FNA) biopsy of the lesion, which was diagnostic of ALHE (Fig. ). Our following surgical attempt included complete mass resection. Despite surgery and postsurgical treatment with topical steroid creams, the lesion recurred. Consequently, we started the patient on intralesional prednisolone twice a month and topical 0.1\% tacrolimus ointment twice daily. This latter regimen seemed to slightly control the lesion’s growth, causing a limited regression in size after 4 months of treatment (Fig. ). A timeline of the patient’s case can be seen in (Fig. ). Our patient’s parents reported her full adherence to treatment. The patient herself reported decreased quality of life and impaired social interactions due to the disfiguring lesions. She also reported marked fear and distress because of the ineffectiveness of multiple therapeutic regimens and surgeries. The patient’s family reported severe financial burden due to the high costs of the treatments. The patient was evicted from the rented house due to a seires of late rent payments recently.\\
   \bottomrule
   \end{tabularx}
   \caption{Sample Patient Notes Across Different Training Datasets}
   \label{tab:note_sample_table}
\end{table}

\begin{table}[htbp]
\centering
\caption{Pipelines for Augmenters and Annotators}
\begin{tabular}{c}
\toprule
\textbf{Algorithm 1: Pipeline for Each Augmenter} \\

\begin{minipage}{0.95\linewidth}
\begin{algorithm}[H]
\begin{algorithmic}[1]
\Require Definition of SDoH category $D$, Raw note $N$
\Ensure Optimized prompt, Verified augmented notes
\State \textbf{Model:} LLM
\State $input\_prompt \gets D$ \Comment{Initialize with category definition}
\While{True}
    \State $augmented\_notes \gets \{\}$
    \State $output \gets$ LLM($input\_prompt$) \Comment{Generate notes based on $D$ and $N$}
    \For{each note $n$ in $output$}
        \State $verification, feedback \gets$ HumanVerify($n$)
        \If{$verification == \text{True}$}
            \State $augmented\_notes \gets augmented\_notes \cup \{n\}$
        \Else
            \State Save $feedback$
        \EndIf
    \EndFor
    \State $accuracy \gets \frac{\text{Number of True verifications}}{\text{Total number of notes}}$
    \If{$accuracy \geq 0.90$}
        \State \textbf{break}
    \Else
        \State $input\_prompt \gets$ OptimizePrompt($input\_prompt$, $feedback$)
    \EndIf
\EndWhile
\State \Return $input\_prompt$, $augmented\_notes$
\end{algorithmic}
\end{algorithm}
\label{alg:augmenter_pipeline}
\end{minipage}
\\

\textbf{Algorithm 2: Pipeline for Annotators} \\

\begin{minipage}{0.95\linewidth}
\begin{algorithm}[H]
\begin{algorithmic}[1]
\Require Training set $(notes, labels)$, Development set $(notes, labels)$
\Ensure Trained DSPy models, Classification of SDoH categories
\State \textbf{Model:} DSPy
\State \textbf{Step 1: Binary Classification}
\State $trainset_1 \gets (notes, labels)$ \Comment{Eviction-related (Yes/No)}
\State $model_1 \gets$ TrainDSPy($trainset_1$)
\State $accuracy_1 \gets$ ValidateModel($model_1$, $devset$)

\State \textbf{Step 2: Eviction Multi-Class}
\State $trainset_2 \gets (notes, labels)$ \Comment{7 eviction-related classes}
\State $model_2 \gets$ TrainDSPy($trainset_2$)
\State $accuracy_2 \gets$ ValidateModel($model_2$, $devset$)

\State \textbf{Step 3: Non-Eviction Multi-Class}
\State $trainset_3 \gets (notes, labels)$ \Comment{7 non-eviction classes}
\State $model_3 \gets$ TrainDSPy($trainset_3$)
\State $accuracy_3 \gets$ ValidateModel($model_3$, $devset$)

\State \Return $model_1, model_2, model_3$
\end{algorithmic}
\end{algorithm}
\label{alg:annotate_pipeline}
\end{minipage}
\\

\end{tabular}
\end{table}

\begin{table}[htbp]
    \centering
    \scriptsize 
    \caption{Keywords}
    \label{tab:keywords}
    \begin{tabular}{p{3.2cm} p{13.5cm}} 
        \toprule
        \textbf{SDoH} & \textbf{Keywords} \\  
        \midrule
        t1\_Homelessness & homeless, homelessness, shelter, transitional housing, living in car, living on streets, couch surfing, lacks housing, no fixed residence, no permanent home \\  
        t1\_InadequateHousing & inadequate housing, structural issues, structural problems, deficiencies in plumbing, heating problems, no heating, electrical problems, lack of running water, broken toilet, no toilet, no kitchen, cramped apartment, overcrowded, unsafe housing, unsanitary living, polluted environment, lead exposure, toxic exposure, mold \\  
        t1\_LackOfAdequateFood & food insecurity, limited access to food, insufficient food, lacks variety, lacks nutrients, no supermarkets, difficult to access food, unstable food sources, cannot afford food, malnutrition, undernourished, skipping meals, reliance on food assistance \\  
        t2\_FinancialInsecurity & financial insecurity, economic insecurity, financial concerns, financial stress, financial burden, affordability issues, rising living costs, difficulty covering expenses, budget difficulties, lacks financial literacy, no stable income, debt, financial hardship \\  
        t2\_HousingInstability & housing instability, difficulty paying rent, frequent moves, multiple moves, families sharing housing \\  
        t2\_MaterialHardship & material hardship, difficulty meeting basic needs, utilities cut off, cannot afford clothing, winter coats, cannot afford school supplies, cannot afford health activities, limited resources for essentials, unable to afford medications, basic needs not met, lacks essential household items \\  
        t2\_TransportationInsecurity & transportation insecurity, lack of transportation, no public transportation, transportation issues, cannot afford transportation fare, long walking distances, inaccessible transportation, no car, unreliable transportation, limited mobility, transportation costs prohibitive \\
        \bottomrule
    \end{tabular}
\end{table}

\begin{table}[htbp]
    \centering
    \scriptsize 
    \caption{Error Analysis - Case 1-3}
    \label{tab:case_study_table}
    \begin{tabular}{p{0.8cm} p{14.5cm}} 
        \toprule
        \textbf{Case} & \textbf{Analysis}\\  
        \midrule
        Case 1 & \textbf{Physician Note}: The patient, a retired pathologist, experienced eviction due to ongoing financial strain as his wife battles a chronic illness. Despite his extensive career, their medical expenses have caused significant challenges, so he resides in a group home at [**Hospital1 3494**] now.\newline
        
        \textbf{Ground Truth}: t3\_Eviction\_present\_history\newline \textbf{Prediction}(Llama-3.1-8B-FT): t3\_Eviction\_present\_current\newline
        
        \textbf{Analysis}: This misclassification likely stems from temporal ambiguity in the text. While the eviction explicitly occurred with no specific timeframe mentioned (indicating a historical event), the phrase 'resides in a group home at [Hospital1 3494] now' may have caused the model to associate the current housing situation with ongoing eviction issues. This highlights the challenge in distinguishing between historical events and currently active situations, particularly when the specific timing of the eviction is not mentioned and 'now' appears in the context.\\
        \midrule
        Case 2 & \textbf{Physician Note}: The patient reports a history of housing instability, having faced an eviction process stemming from unpaid rent two months ago. This event has exacerbated her ongoing health issues, such as bilateral lower extremity edema and nocturnal dyspnea, highlighting the interplay between her living situation and overall well-being.\newline 
        
        \textbf{Ground Truth}: t3\_Eviction\_present\_current\newline \textbf{Prediction}(Llama-3.1-8B-FT): t3\_Eviction\_present\_history \newline 
        
        \textbf{Analysis}: In this case, the model misclassified the eviction situation as \texttt{present\_history} rather than \texttt{present\_current}. This error likely stems from multiple factors in the text. First, the explicit mention of "reports a history of housing instability" may have primed the model to categorize the entire situation as historical, despite the more recent "two months ago" timeframe which falls within our one-year threshold for \texttt{present\_current} classification. The lexical cue "history" appears early in the sentence, potentially biasing the model's interpretation of subsequent temporal information. Additionally, phrases like "having faced an eviction process" use past-tense constructions that might further reinforce a historical framing, even though the event occurred recently. This demonstrates how models can be overly influenced by specific terminology like "history" rather than properly processing the actual temporal indicators ("two months ago") in relation to the annotation schema. This challenge highlights the need for models to prioritize explicit timeframes over potentially misleading lexical cues when making temporal classifications according to specific guidelines. \\
        \midrule
        Case 3 & \textbf{Physician Note}: The patient lives with his wife, who has been recently ill, leading to challenges in meal preparation and unintended weight loss. This situation contributes to financial strain,  leading to a current eviction process initiated by their landlord.\newline 
        
        \textbf{Ground Truth}: t3\_Eviction\_pending (rationale='produce the answer. We need to identify any mention of eviction-related social determinants of health (SDoH) in the patient\'s note. The note describes a situation where the patient and his wife are currently undergoing an eviction process initiated by their landlord due to financial strain. This indicates that the eviction process is active and ongoing. Therefore, the most appropriate label for this situation is "t3\_Eviction\_pending".') \newline 
        
        \textbf{Prediction}(Llama-3.2-3B-FT):
        
        \textbf{Training set with Reasoning}: t3\_Eviction\_pending. Output\_Reasoning: determine the appropriate label for the patient's eviction status. The fact states that the patient is currently facing an eviction process initiated by their landlord, indicating that eviction proceedings are ongoing. Since the eviction is not yet completed and the patient is still residing in the property while dealing with financial strain, we categorize this situation as "pending." 
        
        \textbf{Training set without Reasoning}: t3\_Eviction\_present\_current \newline 
        
        \textbf{Analysis}: This case offers valuable insights into how the presence of reasoning in training data affects both prediction accuracy and reasoning quality. When trained with reasoning annotations, the Llama-3.2-3B-FT model correctly identifies the eviction status as "t3\_Eviction\_pending." However, when the same model is trained without reasoning, it incorrectly classifies the situation as "t3\_Eviction\_present\_current," failing to distinguish between an active ongoing eviction process (pending) and a completed recent eviction (present\_current). Comparing the reasoning produced by the model trained with reasoning to the ground truth reasoning, we observe important similarities and differences. Both correctly identify the key evidence: "eviction process initiated by their landlord." Both reasoning paths also correctly emphasize that this indicates an ongoing, active eviction situation rather than a completed one. However, the model's reasoning is more concise and adds the specific detail that "the patient is still residing in the property," which is implied but not explicitly stated in the ground truth reasoning. \\ 
        \bottomrule
    \end{tabular}
\end{table}

\begin{table}[htbp]
    \centering
    \scriptsize 
    \caption{Error Analysis - Case 4-6}
    \label{tab:case_study_table_2}
    \begin{tabular}{p{0.8cm} p{14.5cm}} 
        \toprule
        \textbf{Case} & \textbf{Analysis}\\  
        \midrule
        Case 4 & \textbf{Physician Note}: The individual does not currently use tobacco and has not faced eviction. He drinks up to 10 beers every weekend.  He has used cocaine several times a week for the past several years, but has not used any for a month. She faces potential housing instability after mutually agreeing to terminate her lease. Previously retired and supporting troubled youth, she now navigates social pressures and community resources while managing her lifestyle choices, including alcohol consumption and past smoking history, which may impact her wellbeing.\newline
        
        \textbf{Ground Truth}: t3\_Eviction\_mr\_current (rationale='produce the answer. We need to identify any mention of eviction-related social determinants of health (SDoH) in the individual\'s note. The note describes a situation where the individual has mutually agreed to terminate her lease, which indicates a mutual rescission of the lease. This suggests that the eviction process is not adversarial but rather a mutual agreement to end the lease. Therefore, the most appropriate label for this situation is "t3\_Eviction\_mr\_current" as the mutual rescission is happening currently.') \newline
        
        \textbf{Prediction}(Llama-3.1-8B-FT): t3\_Eviction\_mr\_current (Reasoning: determine the appropriate label for the patient's eviction status. The individual has mutually agreed to terminate her lease, indicating that she is currently in a situation where she is not facing eviction in the traditional sense, but rather has made a decision to vacate the property. This mutual rescission of the lease suggests that she is actively navigating her housing situation, but it does not indicate an eviction that has already occurred or is currently in progress. Therefore, we categorize this situation as a mutual rescission of the lease.) \newline 
        
        \textbf{GPT-as-judge}: False. The reasoning provided does not fully align with the ground truth. The ground truth clearly states that the situation should be labeled as 't3\_Eviction\_mr\_current' because the mutual rescission of the lease is happening currently. However, the reasoning suggests that the individual is not facing eviction in the traditional sense and does not indicate an eviction that has already occurred or is currently in progress. This implies a misunderstanding of the label 't3\_Eviction\_mr\_current,' which is meant to capture the current mutual rescission of the lease. The reasoning fails to recognize that a mutual rescission is a form of eviction-related SDoH and should be labeled as such. Therefore, there is a gap in understanding the categorization of the situation as per the ground truth. \\  
        \midrule
        Case 5 & \textbf{Physician Note}: The patient lives at home with parents, mother is ESRD on HD and father cares for pt. They are in process of eviction because the landlord issues a notice due to missed rent payments stemming from financial strain. \newline
        
        \textbf{Ground Truth}: t3\_Eviction\_pending (rationale='produce the answer. We need to identify any mention of eviction-related social determinants of health (SDoH) in the patient\'s note. The note describes a situation where the patient and their family are in the process of eviction due to missed rent payments caused by financial strain. This indicates that eviction proceedings have started but are not yet completed. Therefore, the most appropriate label for this situation is "t3\_Eviction\_pending".') \newline
        
        \textbf{Prediction}(Llama-3.1-8B-FT): t3\_Eviction\_pending (Reasoning: produce the answer. The facts indicate that the patient is currently facing eviction due to missed rent payments, which have resulted in a notice from the landlord. Since the eviction process has not yet been completed, but is in progress, the status is "pending." The timeframe is not specified as occurring within the current year, so we will label it as "history." Therefore, the appropriate label will reflect that the eviction is not yet finalized, but is actively being pursued.) \newline 
        
        \textbf{GPT-as-judge}: False. The reasoning provided does not logically support the ground truth. The ground truth clearly identifies the situation as 't3\_Eviction\_pending' based on the fact that eviction proceedings have started but are not yet completed. However, the reasoning introduces an additional and incorrect element by labeling the situation as 'history' due to unspecified timing. This contradicts the ground truth, which does not consider the timeframe as a factor for labeling. The correct label should solely reflect the current status of the eviction process, which is 'pending,' without introducing the notion of 'history.' Therefore, the reasoning is incorrect as it deviates from the ground truth by adding an unnecessary and unsupported condition.\\  
        \midrule
        Case 6 & \textbf{Physician Note}: Social History: Lives alone with PCA 20 hours per day, The individual, recently  faces financial strain, increasing the eviction from her rental home due to unpaid rent. Fountuantly her daughter lives nearby, so she moved to her daughter's home. \newline
        
        \textbf{Ground Truth}: t3\_Eviction\_present\_current (rationale='produce the answer. We need to identify any mention of eviction-related social determinants of health (SDoH) in the patient\'s social history. The note indicates that the individual recently faced financial strain, leading to an eviction from her rental home due to unpaid rent. She moved to her daughter\'s home, which suggests that the eviction process was completed, and she found an alternative living arrangement. Therefore, the most appropriate label for this situation is "t3\_Eviction\_present\_current") \newline
        
        \textbf{Prediction}(Llama-3.1-8B-FT): t3\_Eviction\_present\_current (Reasoning: determine the appropriate label for the patient's eviction status. The fact states that the individual is currently facing financial strain and has not paid rent, which has led to an eviction notice. Since the eviction process has not yet been completed but is in progress, we categorize this situation as "present." Additionally, the individual is currently living with her daughter, indicating a temporary housing solution, but the ongoing eviction status suggests that the situation is still precarious.) \newline 
        
        \textbf{GPT-as-judge}: False. The reasoning provided does not logically support the ground truth. The ground truth indicates that the eviction process was completed, as the individual moved to her daughter's home after being evicted due to unpaid rent. This situation is labeled as 't3\_Eviction\_present\_current,' which implies that the eviction has already occurred and the individual is currently dealing with its consequences. However, the reasoning suggests that the eviction process is still in progress and categorizes the situation as 'present,' which contradicts the ground truth. The reasoning fails to acknowledge that the eviction has been completed and misinterprets the current living situation as temporary rather than a result of completed eviction. \\
        \bottomrule
    \end{tabular}
\end{table}

\end{document}